\documentclass{article} 
\usepackage[preprint]{neurips_2026}
\workshoptitle{AI for Chip Design}

\usepackage{amsmath,amsfonts,bm}

\def\eqref#1{equation~\ref{#1}}

\def\1{\bm{1}}

\DeclareMathAlphabet{\mathsfit}{\encodingdefault}{\sfdefault}{m}{sl}
\SetMathAlphabet{\mathsfit}{bold}{\encodingdefault}{\sfdefault}{bx}{n}

\usepackage{amssymb}

\usepackage{svg} 
\usepackage{hyperref}
\usepackage{url}
\usepackage{graphicx}
\usepackage{longtable}
\usepackage{tikz}
\usetikzlibrary{calc,positioning}
\usepackage{booktabs}
\usepackage[table]{xcolor}
\usepackage[most]{tcolorbox}
\usepackage{listings}
\usepackage{colortbl}
\usepackage{multirow}
\usepackage{placeins}
\usepackage{float}
\usepackage{wrapfig}
\usepackage{algorithm}
\usepackage{algpseudocode}
\usepackage{tabularx}
\newcommand{\edaagent}{\textit{EDA agent}}
\newcommand{\chipmemagent}{\textit{EDA agent + ChipMEM}}
\definecolor{tableblue}{RGB}{0,114,206}
\definecolor{appendixblue}{RGB}{76,96,156}
\definecolor{taskppablue}{RGB}{0,114,206}
\definecolor{tasksimteal}{RGB}{0,125,128}
\definecolor{taskformalpurple}{RGB}{126,76,160}
\newcommand{\grade}[2]{\cellcolor{tableblue!#1}#2}
\newcommand{\gradecenter}[2]{\cellcolor{tableblue!#1}\raisebox{0.65ex}[0pt][0pt]{#2}}
\definecolor{proceduralbg}{RGB}{226,242,239}
\definecolor{statisticalbg}{RGB}{252,235,216}
\definecolor{promptcream}{RGB}{252,247,238}
\newtcblisting{promptbox}[1]{
  enhanced,
  listing only,
  title={#1},
  colback=promptcream,
  colbacktitle=promptcream,
  colframe=black,
  coltitle=black,
  fonttitle=\bfseries\rmfamily,
  boxrule=0.8pt,
  titlerule=0.6pt,
  arc=2mm,
  outer arc=2mm,
  left=2mm,
  right=2mm,
  top=1mm,
  bottom=1mm,
  before skip=6pt,
  after skip=8pt,
  listing options={
    basicstyle=\small\rmfamily,
    breaklines=true,
    columns=fullflexible,
    keepspaces=true,
    showstringspaces=false
  }
}
\newcommand{\memoryterm}[2]{{\setlength{\fboxsep}{1.2pt}\colorbox{#1}{\strut #2}}}
\newcommand{\proceduralmemory}{\memoryterm{proceduralbg}{procedural memory}}
\newcommand{\statisticalmemory}{\memoryterm{statisticalbg}{statistical memory}}
\newcommand{\ProceduralMemory}{\memoryterm{proceduralbg}{Procedural memory}}
\newcommand{\StatisticalMemory}{\memoryterm{statisticalbg}{Statistical memory}}

\title{ChipMEM: Verification-Grounded Memory for EDA Agents}

\author{%
\makebox[0pt][c]{%
\large
\begin{tabular}{c}
\textbf{Abdulrahman AlRabah}$^{1}$\thanks{Corresponding author: \texttt{alrabah2@illinois.edu}}\hspace{1.25em}
\textbf{Joshua Mabry}$^{2}$\hspace{1.25em}
\textbf{Dilek Hakkani-T\"ur}$^{1}$\hspace{1.25em}
\textbf{Abdussalam Alawini}$^{1}$\\[0.55em]
\textbf{Hamid Shojaei}$^{3}$\hspace{1.75em}
\textbf{Kartik Hegde}$^{3}$\hspace{1.75em}
\textbf{Sandesh Adhikary}$^{3}$\\[0.35em]
\mdseries
$^{1}$University of Illinois Urbana-Champaign
\hspace{1.5em}
$^{2}$NVIDIA
\hspace{1.5em}
$^{3}$Cadence
\end{tabular}%
}%
}

\begin{document}

\maketitle
\enlargethispage{2\baselineskip}

\begin{abstract}
Large language model (LLM)-based agents use Electronic Design Automation (EDA)
tools to generate and revise register-transfer-level (RTL) designs under
synthesis and verification feedback. Recent methods learn from this feedback by
distilling reusable skills from execution traces or by training on rewards
derived from EDA-tools. Both methods are typically evaluated on
the tasks that produced the experience. Repeated access to benchmark feedback
on the same task can reward task-specific revision rather than creating
reusable knowledge that transfers. We introduce ChipMEM, a
verification-grounded memory layer for EDA agents. It combines cross-task
procedural memory with within-trajectory statistical guidance. Its procedural
component distills and stores a skill only after it passes synthesis,
simulation, or formal checks, rather than relying on model self-assessments. A
Bayesian component maintains hierarchical Beta estimates over tool-call
outcomes and ranks recovery strategies that succeeded under comparable errors.
A common adapter applies the same memory interface to RTL optimization and
testbench-generation agents while preserving each
domain's tools and acceptance criteria. We measure performance on training
tasks and evaluate whether learned skills transfer to unseen tasks. On RTLRewriter-Bench, under matched model and tool settings, ChipMEM
produces equivalence-passing outputs on 39/54 scored designs versus 35/54
without memory; on the 49-design short suite, mean area improvement is
8.69\% versus 5.66\%. On held-out CVDP tasks, ChipMEM with a frozen
procedural library achieves 20/20 accepted outcomes versus 18/20 without
memory in a single evaluation per setting.
\end{abstract}

\begin{figure}[H]
    \centering
    \vspace{-2mm}
    \includegraphics[width=0.75\textwidth,trim=7 120 49 98,clip]{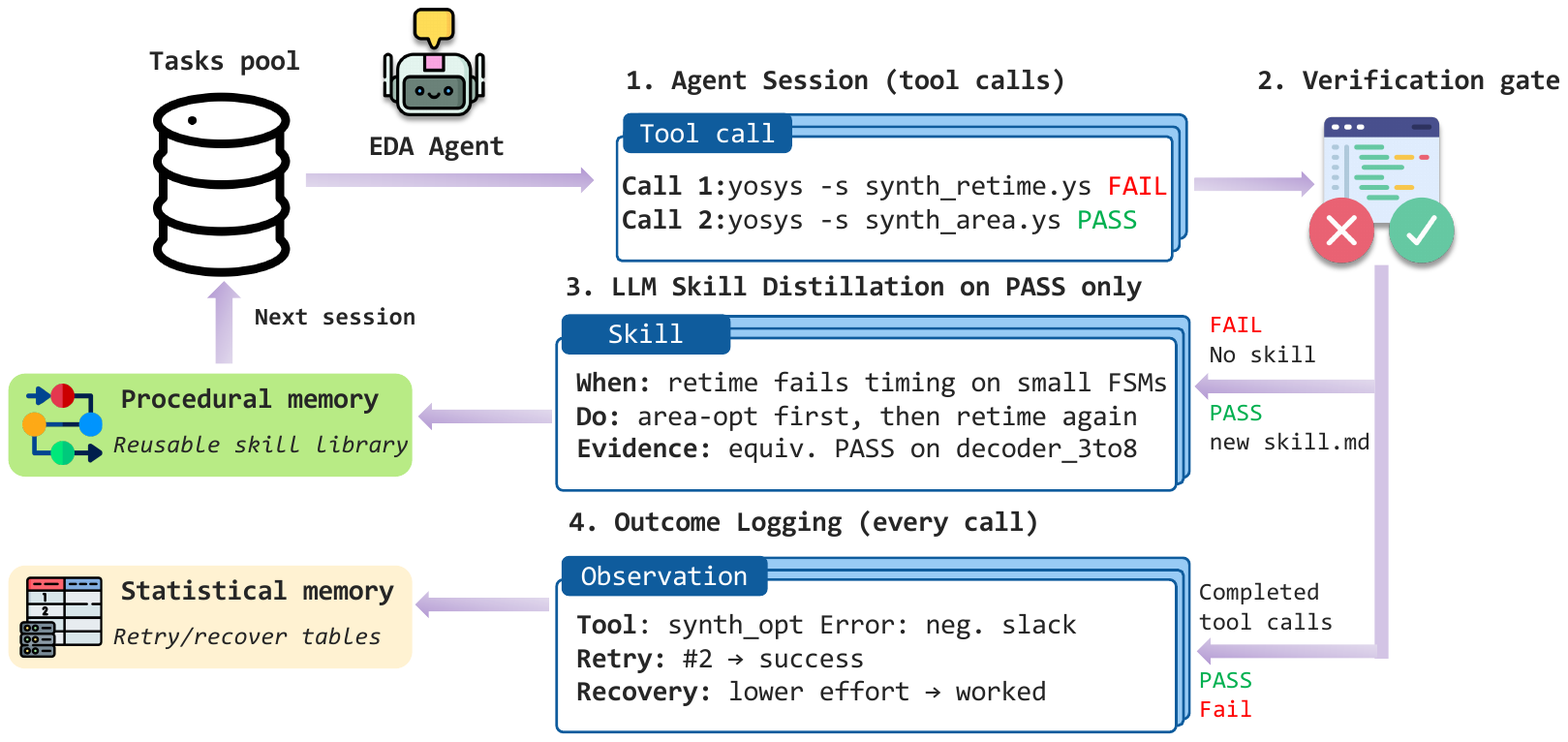}
    \vspace{-1mm}
    \caption{ChipMEM's seed data come from EDA-agent sessions evaluated by domain tools. The verification gate stores a reusable procedural skill only after a passing final artifact, while statistical memory records every completed tool-call outcome to support retry and recovery decisions in later actions and sessions. Figure \ref{fig:memory_architecture} shows the full system architecture, including skill retrieval and use.}
    \vspace{-2mm}
    \label{fig:data-curation}
\end{figure}

\section{INTRODUCTION}
\vspace{-3mm}
Large language models now support several stages of chip design and verification. They can generate register transfer level (RTL) code, revise hardware descriptions, and use feedback from design tools to correct their outputs \citep{liu2023verilogeval,lu2024rtllm,pinckney2025comprehensive,yu2026agentic}. Recent systems also equip agents with reusable skills distilled from prior tool guided sessions \citep{du2026trace2skill,fang2026dr}. This approach reduces the need to encode every lesson in a prompt. Continual learning then depends on whether the agent can discover reliable skills from verified work and reuse them on later tasks. This problem matters in chip development because early RTL decisions shape later design quality. Choices about bit width, resource sharing, and state encoding affect power, performance, and area, yet many useful changes require engineers to restructure the RTL source instead of relying on synthesis alone \citep{lu2026new,yao2024rtlrewriter}. The revised RTL must then pass simulation and equivalence checks before it can move forward. These iterations are slow, and the lessons learned often remain with individual engineers instead of becoming reusable knowledge. Automation is therefore essential.

These challenges have motivated a rapidly growing body of research on large language models for hardware design. Such models can reason over hardware descriptions, propose candidate rewrites in seconds, and iteratively refine them under feedback from EDA tools. Existing efforts have explored both parametric approaches — adapting model weights through supervised fine-tuning or reinforcement learning \citep{chen2026chipseek,shi2025earl,zhou2026alpha} — and nonparametric approaches that equip frozen models with curated skills, search strategies, or tool-in-the-loop feedback \citep{arnold2026rtlscout,ping2026poet}.

Prior work suggests that agents can self-improve by retaining skills, reflections, and strategies from earlier trajectories \citep{wang2023voyager,shinn2023reflexion,ouyang2026reasoningbank}. Recent EDA systems have begun incorporating similar forms of tool-grounded experience. However, these methods are typically evaluated on the same task or task family that produced the memory, leaving unresolved whether stored skills remain relevant \textit{across} tasks and \textit{unseen} designs. Such memory is useful if the agent is faced with the same task again, but the agent must once again start from scratch on brand new tasks. Moreover, naively exposing the agent to memories cultivated from dissimilar tasks has limited utility and may even be counterproductive; the optimal skills for one design may actively diminish performance on a different design. This gap motivates a memory layer whose contents are certified by EDA tools and whose value is measured through transfer beyond the tasks that produced them.
\begin{figure}[t!]
    \centering
    \includegraphics[width=\textwidth,trim=10 55 10 70,clip]{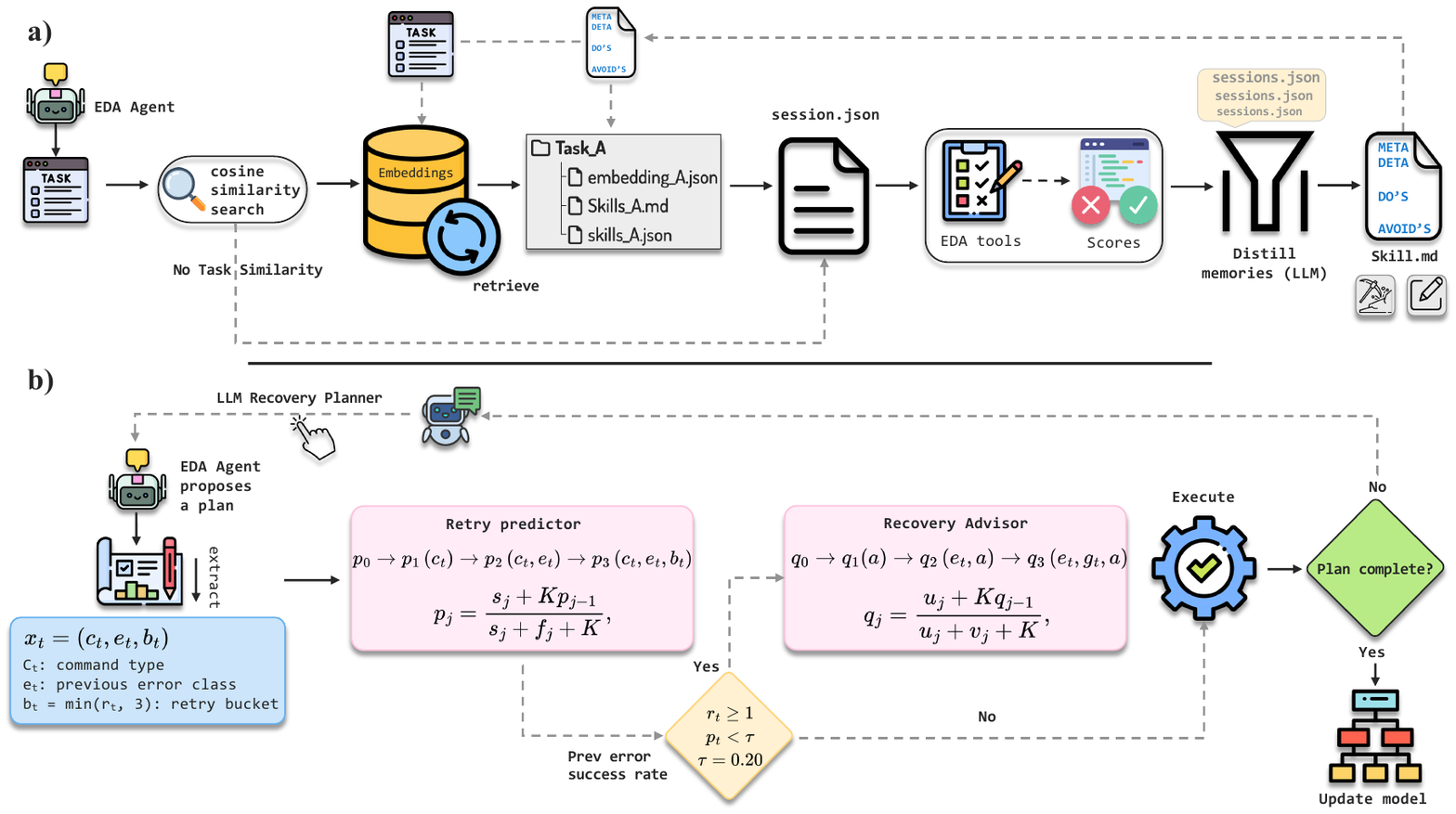}
    \caption{ChipMEM workflow. (a) Procedural memory retrieves skills by task similarity, supplies them to the EDA agent, and stores a distilled skill only after a verified \textsc{pass}. (b) Statistical memory updates from each completed tool call, estimates retry success, and invokes a recovery advisor and bounded LLM plan when the estimate falls below $\tau=0.20$.}
    \label{fig:memory_architecture}
\end{figure}

To address this gap, we propose ChipMEM\footnote{Code: \url{https://github.com/aalrabah/ChipMEM.git}}, a verification-grounded memory layer for cross-task transfer that combines \proceduralmemory{} and \statisticalmemory{}. The \proceduralmemory{} captures complete agent trajectories and distills them into reusable skills to be retrieved for subsequent tasks. These skills are collected and updated according to outcomes from verifiable evaluators for specific tasks, e.g. EDA tools for synthesis and verification. A successful strategy may create a new skill or refine an existing one, while a verified failure may be retained as guidance against repeating an ineffective action. Moreover, the skills are indexed by an embedding derived from the design RTL. Given a new task, similarities with respect to these embeddings are used to retrieve the skills most likely to be related to and beneficial for the current task; the central hypothesis being that similar designs will likely have transferable skills. The \statisticalmemory{} complements these trajectory-derived lessons with a Bayesian model for continual learning through current and historical successes and failures, estimating the probability that a tool strategy will succeed in the current execution context and advising the agent when an unchanged retry is unlikely to be productive. Instead of solely relying on the agent's intrinsic reasoning to identify the right tools, we equip it with a ``data analyst" that produces advice grounded in statistics collected from prior rollouts. With statistical memory, the agent is thus able to begin a session with an informed prior over success likelihoods of tools, and can update this prior with each tool execution.

ChipMEM combines the procedural and statistical memories into a single memory layer through a common agent adapter that records the task representation, retrieved context, execution trajectory, returned artifacts, and domain-specific acceptance criterion, allowing both memory components to support EDA agents without requiring those agents to share the same tools or scoring procedure. This design allows procedural memory to evolve across a sequence of tasks or remain frozen for evaluation on unseen designs, while statistical memory summarizes execution outcomes and provides probabilistic guidance during tool use. 

We summarize our primary contributions as follows:
\begin{enumerate}
    \item ChipMEM combines procedural memory that extracts reusable skills from tool-scored executions with Bayesian statistical memory that accumulates tool-level successes and failures to guide retries and recovery.

    \item ChipMEM provides an agent-agnostic adapter between an EDA agent, its tools, and its evaluation harness. The adapter supplies retrieved skills and recovery guidance, records execution trajectories, and returns artifacts for domain-specific evaluation without modifying the underlying agent or tools.

    \item We evaluate ChipMEM against memory-off baselines across multiple EDA tasks and benchmarks, examining verified outcomes, interaction cost, and the effects of each memory component.

    \item We separately evaluate transfer to unseen tasks using a fixed, read-only procedural library, testing whether prior experience remains useful without adding new skills during evaluation.
\end{enumerate}

We evaluate ChipMEM on the RTL-OPT \citep{lu2026new} and RTLRewriter \citep{yao2024rtlrewriter} benchmarks for PPA optimization, and on CVDP tasks for testbench generation \citep{pinckney2025comprehensive}. The evaluation tests whether skills distilled from verified trajectories transfer to later tasks and unseen designs, and whether accumulated tool outcomes improve decisions during execution.


\section{RELATED WORK}
\paragraph{Skill memory and design-aware retrieval.}
There is a growing body of work that equips frozen agents with procedural memory, storing experience from past episodes as reusable artifacts and retrieving it to guide later tasks. Voyager \citep{wang2023voyager} builds an executable skill library from successful episodes, Reflexion \citep{shinn2023reflexion} turns execution feedback into verbal episodic memory, ReasoningBank \citep{ouyang2025reasoningbank} distills transferable strategies from both successful and failed trajectories, and SkillOpt \citep{yang2026skillopt} refines a skill document over a frozen model without weight updates. In chip design, however, the relevance of a stored strategy depends on the target specification, RTL structure, tool context, and failure mode, so a skill that helps one design may be ineffective or harmful on another. This makes retrieval as important as skill extraction and motivates evaluating memory the task that produced it. ChipMEM extends this direction by integrating trajectory-distilled procedural memory for cross-task transfer with Bayesian statistical memory for within-session guidance, while grounding both in outcomes produced by EDA tools.

\paragraph{Tool-grounded RTL optimization.}
RTL-OPT \citep{lu2026new} and RTLRewriter \citep{yao2024rtlrewriter} evaluate LLM-driven RTL optimization under both implementation-quality and functional-equivalence criteria. Recent methods then fold EDA feedback into optimization in different ways. Reinforcement learning drives a gated toolchain reward into model weights \citep{chen2026chipseek}. Evolutionary and agentic search explore candidate rewrites under synthesis feedback \citep{ping2026poet,arnold2026rtlscout,hsin2026evolve}. Frozen agents instead evolve reusable skills from execution traces \citep{du2026trace2skill,fang2026dr,wang2026veriagent}, and test-time training adapts weights to a single design and discards them once it changes \citep{zhou2026alpha}. These methods improve how an agent searches or adapts during optimization, whereas ChipMEM turns verified outcomes into reusable procedural and statistical memory that transfers to later tasks and different EDA agents.

\paragraph{LLMs for RTL design and evaluation.}
VerilogEval \citep{liu2023verilogeval}, RTLLM \citep{lu2024rtllm}, and CVDP \citep{pinckney2025comprehensive} provide established benchmarks for RTL generation, completion, and verification. Building on these efforts, ChipMEM extends the evaluation beyond isolated task success to examine whether tool-verified experience remains useful across later tasks and unseen designs, and whether historical outcomes can improve decisions during tool-guided execution.

\section{ChipMEM}
ChipMEM is an agent-agnostic layer that converts tool-verified execution traces into two complementary memories: \proceduralmemory{} preserves reusable strategies across tasks, while \statisticalmemory{} learns action-level success and recovery patterns within and across sessions. Both components augment the agent context without changing the underlying EDA agent, tools, or domain-specific acceptance gate. Figure~\ref{fig:memory_architecture} summarizes this dual-memory workflow.

\paragraph{\ProceduralMemory{}.}
We represent the procedural memory library as key-value map $\mathcal{M}^{\mathrm{proc}}=\{k_i\mapsto(v_i,m_i,h_i)\}_{i=1}^{M}$. For a given task, the value tuple $(v_i,m_i,h_i)$ consists of the the complete skill file $v_i$, metadata $m_i$ for its domain, task, provenance, and verifier outcomes $h_i$. For the $i$-th task, the key $k_i=\phi(z_i) \in \mathbb{R}^d$ is the task embedding obtained by applying an embedding model $\phi$ to a pre-defined task-artifact $z_i$. For instance, in our experiments, $z_i$ corresponds to the top RTL file for RTL optimization and the task-definition \texttt{PROMPT.md} file for CVDP tasks. This task definition excludes agent instructions, retrieved skills, tool history, generated files, tests, and scoring artifacts.

The library starts empty and grows across $N_{\mathrm{seed}}$ tasks run for $R$ rounds. For the $n$-th task, we retrieve procedural-memory as value-tuples with keys exhibiting the largest similarity to the current task's embedding $q_n$, i.e., 
\[
\mathcal{R}_n=\left\{(v_i,m_i,h_i)\;\middle|\;i\in\operatorname{TopK}\left[\mathrm{sim}(q_n,k_j);~\tau_{\mathrm{proc}}\right]\right\}.
\]
Here, $\operatorname{TopK}[\texttt{sim}(\cdot);\tau_{\mathrm{proc}}]$ is the list of top-k matches with respect to a similarity function $\texttt{sim}$ and a threshold $\tau_\mathrm{proc}$; only matches with similarities higher than $\tau_\mathrm{proc}$ are returned. We use cosine-similarity over embeddings as our similarity function $\texttt{sim}(q_n,k_j)$. Thus $\mathcal{R}_t$ contains at most $K$ complete memory items with the highest task similarity above a threshold $\tau_{\mathrm{proc}}$. 

Once the $n$-th task is processed, ChipMEM may expand the procedural memory library by adding a skill for the source task if none exist. Skills are added to the library only if the corresponding task execution obtains a verified \textsc{pass} from a verifiable domain-specific evaluation harness, not an LLM judge. We can apply the procedural memory library under two modes: evolving and frozen. While the evolving-mode carries verified additions forward to later tasks, the frozen-mode keeps its initial library read-only throughout.

\paragraph{\StatisticalMemory{}.}
\begingroup
\setlength{\abovedisplayskip}{4pt plus 1pt minus 1pt}
\setlength{\belowdisplayskip}{4pt plus 1pt minus 1pt}
\setlength{\abovedisplayshortskip}{2pt plus 1pt}
\setlength{\belowdisplayshortskip}{3pt plus 1pt minus 1pt}
While procedural memory allows EDA agents to extract high-level strategies to aid their tasks, agents can also benefit from the more structured and narrower-scoped skill of tool-failure-recovery. EDA agents often repeat failed tool calls without reusing recoveries that have succeeded in earlier trajectories. Since EDA-tools can have particularly high latency, repeated tool-call failures can lead to significant delays and wasted token-consumption. Thus, we propose statistical memory that maintains a prior belief over tool-call success based on statistics collected from prior runs, updates the prior based on the current trajectory, and provides advice on tool-retry-success and failure-recovery strategies by learning on a tool-step granularity. The statistical memory component is comprised of two binomial distribution models -- the retry predictor and the recovery predictor -- encoding the success likelihood of tool-calls and their corresponding failure-recovery actions. As illustrated in Figure~\ref{fig:memory_architecture}, the retry predictor informs the agent whether its proposed tool-call is likely to succeed, and the recovery predictor provides the most-successful strategy to recover from that failure.

At the $t$-th turn in the execution trajectory of an agent, let $c_t$ be tool-call submitted by the agent; further assume that this tool has previously been called $r_t$ times in the same session with the last call resulting in error $e_t$. If the previous call for $c_t$ succeeded, $e_t$ is set to $\texttt{None}$; otherwise it is set to the specific error class encountered. For example, in our experiments with the PPA agent, such error classes include \texttt{syntax-error}, \texttt{unknown-module}, \texttt{liberty-missing}, \texttt{timing-fail}, and \texttt{rc-nonzero}. This provides us with statistics on the likelihood of a tool's success. Following a tool-call error, we also record the action $a_t$ taken by the agent, and whether the $a_t$ resulted in a successful recovery. For each error class $e_t$, we thus collect a table of associated recovery actions and their corresponding likelihoods of success. 

We define the context of the retry-model via the tuple $x_t=(c_t,e_t,b_t)$, where $b_t = \min(r_t,3)$. We set $b_t=\min(r_t,3)$, so its values are $0$, $1$, $2$, and $3$, with $3$ grouping three or more earlier calls. This saturation prevents sparse evidence from being split across many large retry counts. Starting from the global Beta$(1,1)$ posterior $p_0$, the model adds one conditioning variable at each of three levels. Level $p_1$ conditions on command type, $p_2$ adds $1$-step history of the previous error class, and $p_3$ adds prior history:
$p_0\rightarrow p_1(c_t)\rightarrow p_2(c_t,e_t)\rightarrow p_3(c_t,e_t,b_t)$, with
\begin{equation}
    p_j=\frac{s_j+Kp_{j-1}}{s_j+f_j+K},
    \qquad j\in\{1,2,3\},
\end{equation}
where $s_j$ and $f_j$ are prior successful and failed calls at level $j$, and the parameter $K$ essentially controls the weight placed on prior history; a lower value of $K$ makes the model more myopic. In our experiments, we use $K=8$. The final estimate is $P_{\mathrm{retry}}(\mathrm{success}\mid c_t,e_t,b_t)=p_3$. After a failed same-type call, a value below $\tau_{\text{stat}}=0.20$ activates the recovery model.

For a candidate recovery strategy $a$ and the current tool stage $g_t$. The recovery hierarchy is
\[
q_0 \rightarrow q_1(a) \rightarrow q_2\left(a, e_t\right) \rightarrow q_3\left(a,e_t,g_t\right).
\]
For example, $q_1(\texttt{inspect\_logs})$ estimates how often inspecting logs has recovered a failure across all errors and stages. At each level,
\begin{equation}
    q_j=\frac{u_j+Kq_{j-1}}{u_j+v_j+K},
    \qquad j\in\{1,2,3\},
\end{equation}
where $u_j$ and $v_j$ are historical successful and failed recoveries, $q_{j-1}$ is the probability inherited from the broader level, and $K=8$. Hence $P_{\mathrm{recovery}}(a\mid e_t,g_t)=q_3$, and candidate strategies are ranked as
\begin{equation}
    \pi(e_t,g_t)=\operatorname{sort}_{a}\,
    P_{\mathrm{recovery}}(a\mid e_t,g_t).
\end{equation}
When activated, the recovery model passes its ranked strategies to an LLM recovery planner, which returns a bounded advisory plan to the agent. The ranking is advisory; each completed tool call atomically persists retry and recovery evidence for the next decision, including the next decision in the same session.
\endgroup

\paragraph{Algorithm.}
Algorithm~\ref{alg:chipmem} summarizes ChipMEM across ordered sessions. Here $D_n$ is the task artifact, $\mathcal{A}$ is the EDA agent, $\mathcal{T}$ is its tool set, $\mathcal{V}$ is the deterministic domain evaluation harness, $q_n$ is the task embedding, $\mathcal{R}_n$ is the retrieved skill set, $H_n$ is the execution trajectory, and $y_n$ is the final harness verdict. \statisticalmemory{} updates after every tool call. Only a passing trajectory can add a skill to \proceduralmemory{}. The algorithm shows evolving mode. Frozen mode starts from a fixed library and skips the skill storage step.

\begin{algorithm}[H]
\caption{\proceduralmemory{} and \statisticalmemory{} across sessions}
\label{alg:chipmem}
\small
\begin{algorithmic}[1]
\Require Tasks $\{D_n\}_{n=1}^{N}$, agent $\mathcal{A}$, EDA tools $\mathcal{T}$, domain evaluation harness $\mathcal{V}$, embedder $\phi$
\State $\mathcal{M}^{\mathrm{proc}} \gets \emptyset$; initialize $\mathcal{M}^{\mathrm{stat}}$ with $\mathrm{Beta}(1,1)$ priors \Comment{procedural library; Bayesian state}
\For{$n = 1,\ldots,N$} \Comment{$n$: session; $N$: total sessions}
    \State $q_n \gets \Call{EmbedTask}{\phi, D_n}$ \Comment{$q_n$: task embedding}
    \State $\mathcal{R}_n \gets \Call{RetrieveSkills}{\mathcal{M}^{\mathrm{proc}}, q_n, \tau_{\mathrm{proc}}}$ \Comment{$\mathcal{R}_n$: top $K$ skill payloads; $\tau_{\mathrm{proc}}$: threshold}
    \State Give $(D_n, \mathcal{R}_n)$ to $\mathcal{A}$; initialize $H_n \gets \emptyset$ \Comment{$H_n$: trajectory}
    \While{session $n$ is active}
        \State $\mathcal{A}$ selects an action and calls a tool in $\mathcal{T}$ \Comment{\textbf{Act}}
        \State Append the action and returned outcome to $H_n$
        \State $\mathcal{M}^{\mathrm{stat}} \gets \Call{BayesUpdate}{\mathcal{M}^{\mathrm{stat}}, \text{latest tool outcome}}$ \Comment{$x_t=(e_t,c_t,b_t)$; retry model}
        \If{a repeated failure has $P_{\mathrm{retry}} < \tau_{\mathrm{B}}$} \Comment{retry probability; nudge threshold}
            \State Advise $\mathcal{A}$ using $\pi(e_t, g_t)$ \Comment{recovery model; $e_t$: error; $g_t$: stage}
        \EndIf
    \EndWhile
    \State $y_n \gets \Call{RunDomainHarness}{\mathcal{V}, D_n, H_n}$ \Comment{$y_n$: final harness verdict}
    \If{$y_n = \textsc{pass}$}
        \State $v_n \gets \Call{DistillProcedures}{H_n}$ \Comment{procedural distiller; $v_n$: skill}
        \State $\mathcal{M}^{\mathrm{proc}} \gets \Call{StoreSkill}{\mathcal{M}^{\mathrm{proc}}, q_n, v_n}$ \Comment{$q_n$: index; $v_n$: skill}
    \Else
        \State Leave $\mathcal{M}^{\mathrm{proc}}$ unchanged
    \EndIf
\EndFor
\end{algorithmic}
\end{algorithm}

\paragraph{Agent adapter and verification.}
ChipMEM connects to each EDA domain through a common adapter. For task $D_n$, the adapter gives $\mathcal{A}$ the original task, retrieved skills $\mathcal{R}_n$, and any Bayesian recovery advice. It exposes the domain tools, records each tool action and result in $H_n$, and returns the final artifact to $\mathcal{V}$. Here $\mathcal{V}$ is a deterministic domain evaluation harness, not an LLM judge. PPA acceptance requires successful synthesis, functional equivalence, and a positive audited improvement. CVDP testbench generation requires the hidden simulation, coverage, and mutation checks to pass. In evolving mode, a verified \textsc{pass} may add a procedural skill. Frozen mode leaves \proceduralmemory{} unchanged. A \textsc{fail} or \textsc{invalid} verdict creates no skill, while \statisticalmemory{} still records completed tool outcomes.

\FloatBarrier
\section{Experimental Setup}

\paragraph{Tasks}
The CVDP study uses 40 testbench tasks, with 20 training tasks and 20 held-out tasks split evenly between CID012 stimulus generation and CID013 checker generation \citep{pinckney2025comprehensive}. RTL-OPT contains 38 design-level RTL optimization tasks \citep{lu2026new}, while RTLRewriter-Bench contains 59 available cases, 54 short and five long \citep{yao2024rtlrewriter}. OpenTitan evaluates ten IP blocks over five attempts per mode \citep{meza2023opentitan}.
Additionally, we also performed evaluations on a custom set of $20$ open-sourced designs (see Appendix for details).

\paragraph{Implementation details.}
We use GPT-5.5 \citep{openai2026gpt55} for RTL-OPT, RTLRewriter-Bench, the custom-design PPA study, and CVDP CID012 and CID013. We use Qwen3.8-27B \citep{qwen2026qwen38} for OpenTitan. \ProceduralMemory{} retrieval uses \texttt{text-embedding-3-large} for the first group and \texttt{Qwen/Qwen3-Embedding-0.6B} for OpenTitan. Reasoning is set to medium for all experiments. OpenTitan uses at most 40 tool steps, a 600 second model timeout, and a 7200 second session limit. Retrieval returns at most two skills at similarity threshold $0.6$. The open model and its embedding model run on an NVIDIA A100 80GB GPU \citep{nvidia2020a100}. Appendix~\ref{app:execution-metrics} reports the detailed resource measurements.

\paragraph{Evaluation metrics.}
Equivalence pass rate and strict gate pass rate are binary measures of functional correctness and complete task acceptance. For PPA, area, power, and timing gains are measured relative to the original design after synthesis and equivalence. Failed, invalid, unproven, unchanged, and nonimproving outputs receive zero gain. RTLRewriter additionally reports geometric mean cell and wire ratios relative to the baseline, where values below one indicate reduction. For CID012 and CID013, task pass rate is the fraction of generated testbenches that pass every hidden Xcelium/IMC test. For ordered passes, cumulative Pass@$k$ is the fraction of tasks solved by pass $k$ and is reported as a sequential curve rather than an independent sampling estimate.

An EDA agent works on one design over many turns. It reads and edits files, calls design tools, inspects failures, and revises its output before the final evaluation. The PPA agent can call synthesis, simulation, timing, and equivalence tools. The simulation agent can compile and run testbenches. The CVDP simulation set contains 20 training tasks and 20 unseen tasks. CID012 stimulus generation and CID013 checker generation each contribute 10 training tasks and 10 unseen tasks. The unseen tasks are disjoint from training. The agent sees the task files and design context, while hidden tests and reference artifacts remain outside its workspace.



\begin{figure}[t!]
\label{fig:ppa_results}
    \centering
    \includegraphics[scale=0.46]{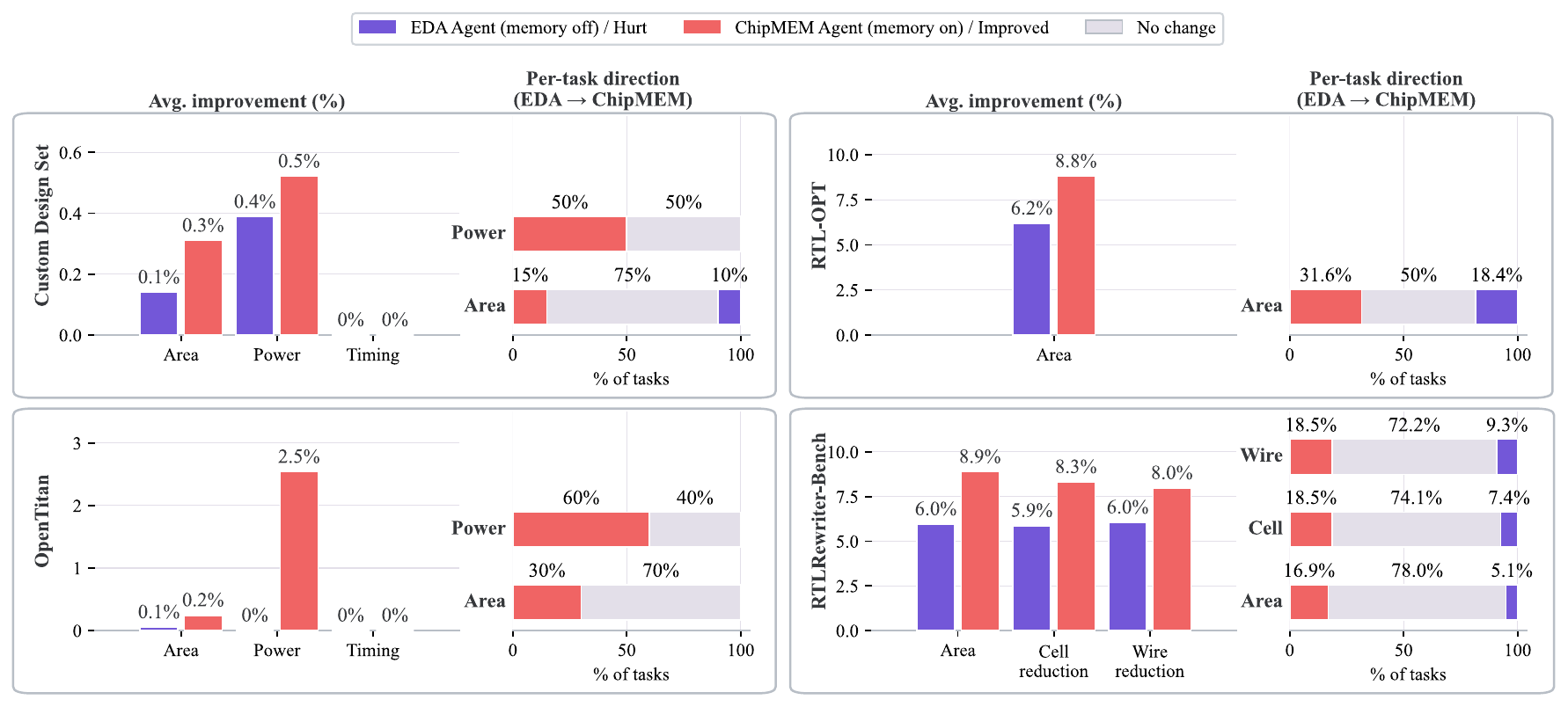}
    \caption{Audited PPA outcomes across the custom-design PPA suite, OpenTitan, RTL-OPT, and RTLRewriter-Bench. Vertical bar graphs compare baseline EDA-Agent with no memory (purple) against ChipMEM (red). Horizontal bars show the per-task direction of change under ChipMEM as improved, unchanged, or hurt. RTLRewriter means use 54 scored cases after excluding five zero-cell cases. For, RTL-OPT we select the best valid result per design across three result sets and reports the strongest demonstrated capability.}
    \label{fig:ppa-area-improvement}
\end{figure}

\paragraph{PPA metric scope.}
The PPA studies use two evaluation scopes. We began with RTL-OPT and RTLRewriter-Bench as first-pass area-optimization studies using Yosys for synthesis and equivalence and ABC for technology mapping \citep{wolf2013yosys,brayton2010abc}. This fast, reproducible loop let us verify that the agent workflow, equivalence gate, and measurement methodology operated correctly before moving to the broader Custom Design Set and OpenTitan flows. The first-pass runs measure mapped area, and RTLRewriter-Bench also records cell and wire counts. Because we did not run a matched characterized downstream power or timing flow for these two benchmarks, we do not report unmeasured power or performance values. The later Custom Design Set and OpenTitan studies report full PPA metrics, including area, power, and timing. Figure~\ref{fig:ppa-area-improvement} presents the first-pass area-optimization and full-PPA results together under their respective measurement scopes.

\paragraph{Baselines and ablations.}
Figure~\ref{fig:ppa-area-improvement} and Appendix Table~\ref{tab:cvdp-simulation-results} compare \edaagent{}\footnote{The EDA Agent is Cadence ChipStack 2.0, a proprietary system actively used in production by leading chip-design companies; all GPT-5.5 experiments were run on this agent.}
with \chipmemagent{}. \edaagent{} is the same domain agent with memory disabled. It receives no retrieved skills, writes no skills, and receives no Bayesian guidance. \chipmemagent{} uses the same model, task, prompt, tools, and budget with memory enabled. In evolving mode, \chipmemagent{} updates the skill library after verified passing tasks. In frozen mode, \chipmemagent{} reads a fixed library during unseen task evaluation. These settings provide the whole-system comparison between Memory OFF and full ChipMEM. Separately, on five RTL-OPT designs, we compare Memory OFF, procedural-only, and Bayesian-only under matched settings, while the procedural-plus-Bayesian loaded-state replay reports combined system capability.
\vspace{-2mm}
\section{Results and Discussion}
\vspace{-1mm}
\paragraph{Cross benchmark interpretation.}
Figure~\ref{fig:ppa-area-improvement} shows that, on the two public benchmarks, ChipMEM increases mean area improvement from 6.18\% to 8.79\% on RTL-OPT and from 5.95\% to 8.88\% on RTLRewriter-Bench. It also raises RTLRewriter-Bench cell and wire reductions from 5.9\%/6.0\% to 8.3\%/8.0\%. The per-design results include 12 area improvements on RTL-OPT and 10 on RTLRewriter-Bench, showing that the gains extend across distinct public RTL optimization tasks. In the full-PPA studies, the Custom Design Set increases mean area and power improvement from 0.14\%/0.39\% to 0.31\%/0.52\%. OpenTitan increases mean area and power improvement from 0.05\%/0.00\% to 0.24\%/2.54\%, while timing remains stable. Table~\ref{tab:ppa-results} further shows that OpenTitan full-PPA gate passes rise from 1/10 to 4/10 with equivalence maintained at 6/10 in both modes. Together, these results show that ChipMEM improves both public area-optimization benchmarks and the broader full-PPA design flows.


\FloatBarrier
\begin{wrapfigure}{r}{0.50\textwidth}
    \vspace{-9mm}
    \centering
    \includegraphics[width=\linewidth]{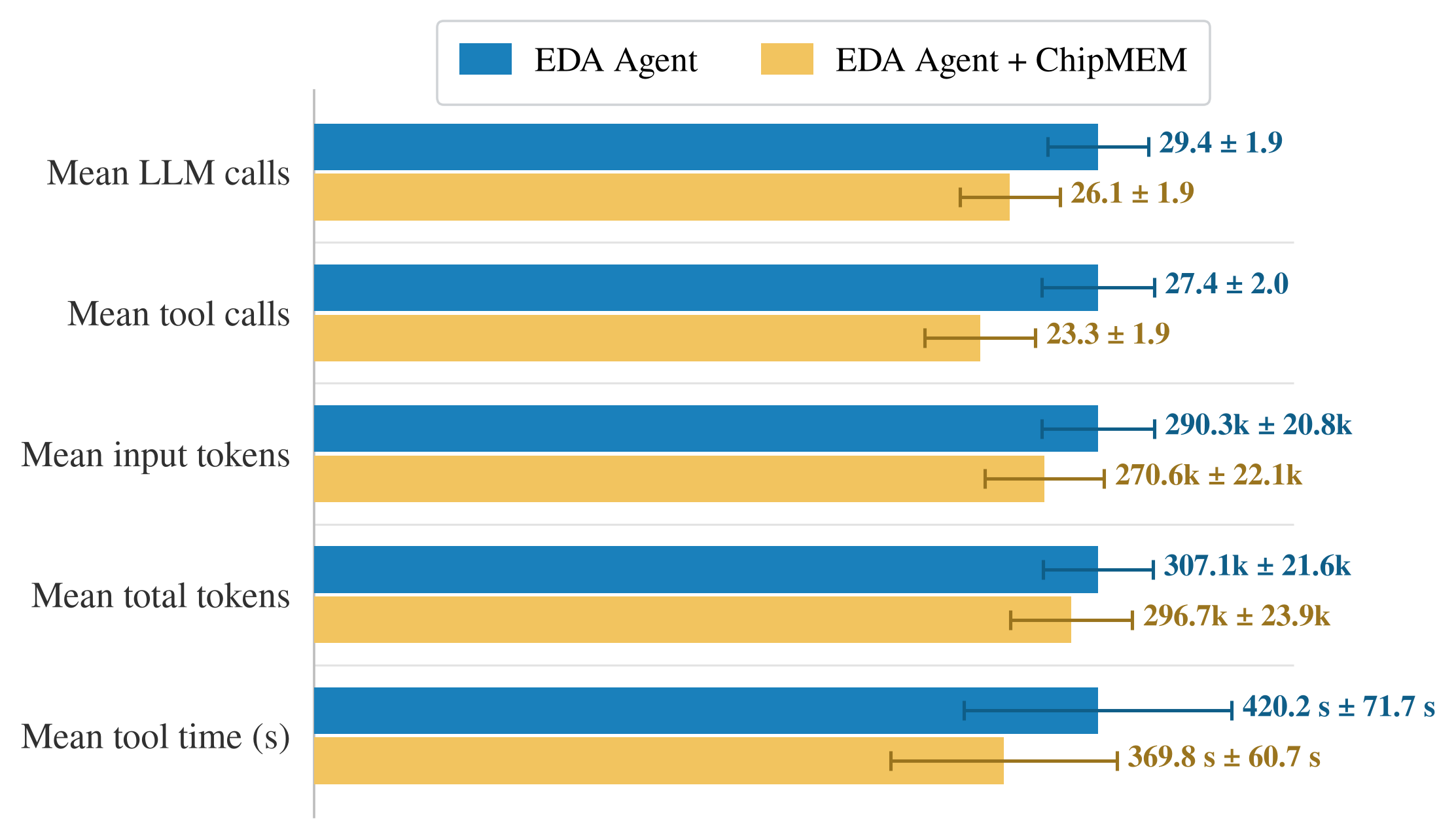}
    \vspace{-4mm}
    \caption{OpenTitan interaction cost over 50 sessions per mode, comprising ten designs and five attempts per design. Bars and labels report mean $\pm$ standard error for LLM calls, tool calls, input and total tokens, and tool time.}
    \label{fig:opentitan-memory-efficiency}
    \vspace{-6mm}
\end{wrapfigure}
\paragraph{OpenTitan interaction cost.}
Figure~\ref{fig:opentitan-memory-efficiency} reports lower means for ChipMEM on numerous resource measures. Mean LLM calls fall from 29.4 to 26.1, mean tool calls from 27.4 to 23.3, and mean tool time from 420.2 to 369.8 seconds, corresponding to reductions of 11.2\%, 15.0\%, and 12.0\%. Input tokens decrease by 6.8\%, and total tokens decrease by 3.4\%. Across the 50 sessions, ChipMEM has lower average LLM calls, tool calls, input and total tokens, and tool time.

\vspace{-3mm}
\paragraph{Memory ablation. }
We evaluate four memory configurations on the same RTL-OPT designs ($n=5$) with matched model, task order, RTL inputs, tools, seed, and execution limits. Table~\ref{tab:rtlopt-memory-ablation} reports strict PASS counts, audited area gain, and interaction cost. Procedural and Bayesian each produce four strict passes, compared with three for Memory OFF. The combined state produces five strict passes, the highest mean audited gain, and the lowest tokens per accepted design. It records the capability of the full system after both memories accumulated prior experience on the same design sequence.

\begin{table}[t!]
\centering
\footnotesize
\setlength{\tabcolsep}{5pt}
\renewcommand{\arraystretch}{1.1}
\caption{Five-design RTL-OPT memory study ($n=5$). Tokens report aggregate input/output tokens. Tool/LLM reports tool calls and successful agent LLM turns. Time is summed session wall time. Best values are bold.}
\label{tab:rtlopt-memory-ablation}
\begin{tabularx}{\textwidth}{@{}l *{6}{>{\centering\arraybackslash}X}@{\hspace{5pt}}}
\toprule
\rowcolor{gray!20}
\multicolumn{7}{c}{\textit{Memory component ablation and loaded state replay ($n=5$)}} \\
\midrule
\textbf{Mode} & \textbf{PASS} & \textbf{Gain} & \shortstack{\textbf{Tokens}\\\textbf{in/out}} & \shortstack{\textbf{Tokens}\\\textbf{/PASS}} & \shortstack{\textbf{Tool}\\\textbf{/LLM}} & \textbf{Time} \\
\midrule
Memory OFF & 3 & 8.41\% & 1.82M/150k & 657k & 183/191 & 111.5m \\
Procedural only & 4 & 10.13\% & 1.67M/96k & 442k & 173/184 & 66.7m \\
Bayesian only & 4 & 8.60\% & \textbf{1.29M}/84k & 344k & 135/147 & 67.7m \\
\textbf{Procedural + Bayesian} & \textbf{5} & \textbf{10.36\%} & 1.49M/\textbf{72k} & \textbf{312k} & \textbf{119/133} & \textbf{49.2m} \\
\bottomrule
\end{tabularx}
\end{table}

\begin{figure}[t!]
  \centering
  \resizebox{\textwidth}{!}{%
\definecolor{pstflow}{RGB}{0,114,206}
\definecolor{pstsetup}{RGB}{126,76,160}
\definecolor{pstrtl}{RGB}{0,125,128}
\definecolor{pstink}{RGB}{36,42,48}
\definecolor{pstsubtle}{RGB}{92,99,106}

\tikzset{
  pstcard/.style={
    draw=#1!55,
    fill=#1!5,
    rounded corners=2.5pt,
    line width=0.55pt,
    minimum width=5.50cm,
    minimum height=4.95cm,
    inner sep=0pt
  },
  psttag/.style={
    fill=#1,
    text=white,
    rounded corners=1.5pt,
    font=\sffamily\bfseries\footnotesize,
    inner xsep=7pt,
    inner ysep=3.5pt
  },
  pstarchetype/.style={
    font=\sffamily\footnotesize,
    text=pstsubtle,
    align=center,
    text width=4.70cm,
    execute at begin node={\hyphenpenalty=10000\relax}
  },
  psttearout/.style={
    draw=#1!52,
    fill=white,
    rounded corners=1.5pt,
    dash pattern=on 2.2pt off 1.6pt,
    line width=0.65pt,
    minimum width=4.92cm,
    minimum height=2.30cm,
    inner sep=0pt
  },
  pstyaml/.style={
    font=\ttfamily\scriptsize,
    text=pstink,
    align=left,
    text width=4.42cm,
    execute at begin node={\hyphenpenalty=10000\relax}
  }
}

\begin{tikzpicture}[x=1cm,y=1cm]
  \node[pstcard=pstflow] (flowcard) at (0,0) {};
  \node[psttag=pstflow,anchor=north] at ([yshift=-0.28cm]flowcard.north)
    {Flow \& Validation};
  \node[pstarchetype,anchor=north] at ([yshift=-0.96cm]flowcard.north)
    {General flow knowledge for establishing a valid synthesis baseline and trustworthy PPA evidence.};
  \node[psttearout=pstflow] (flownote) at ([yshift=-1.08cm]flowcard.center) {};
  \node[pstyaml] at (flownote.center) {\textcolor{pstflow}{skill-name:} sv-cdc-fifo-master\\[2pt]
    \textcolor{pstflow}{skill-summary:}\\
    \hspace*{1em}Use the intended FIFO top\\
    \hspace*{1em}and full hierarchy with every\\
    \hspace*{1em}clock and valid CDC constraint\\
    \hspace*{1em}before interpreting PPA.};

  \node[pstcard=pstsetup,right=0.28cm of flowcard] (setupcard) {};
  \node[psttag=pstsetup,anchor=north] at ([yshift=-0.28cm]setupcard.north)
    {Design Setup};
  \node[pstarchetype,anchor=north] at ([yshift=-0.96cm]setupcard.north)
    {Design-specific knowledge needed to compile and elaborate a particular IP correctly.};
  \node[psttearout=pstsetup] (setupnote) at ([yshift=-1.08cm]setupcard.center) {};
  \node[pstyaml] at (setupnote.center) {\textcolor{pstsetup}{skill-name:} sram-ctrl\\[2pt]
    \textcolor{pstsetup}{skill-summary:}\\
    \hspace*{1em}Preserve SRAM controller\\
    \hspace*{1em}package and source order,\\
    \hspace*{1em}include paths, defines, and\\
    \hspace*{1em}assertion macros; local shims\\
    \hspace*{1em}can leave hierarchy incomplete.};

  \node[pstcard=pstrtl,right=0.28cm of setupcard] (rtlcard) {};
  \node[psttag=pstrtl,anchor=north] at ([yshift=-0.28cm]rtlcard.north)
    {RTL Optimization};
  \node[pstarchetype,anchor=north] at ([yshift=-0.96cm]rtlcard.north)
    {RTL transformation knowledge, including what may change and which behaviors must be preserved.};
  \node[psttearout=pstrtl] (rtlnote) at ([yshift=-1.08cm]rtlcard.center) {};
  \node[pstyaml] at (rtlnote.center) {\textcolor{pstrtl}{skill-name:} priority-encoder\\[2pt]
    \textcolor{pstrtl}{skill-summary:}\\
    \hspace*{1em}Enumerated case or threshold\\
    \hspace*{1em}rewrites increased mapped area\\
    \hspace*{1em}or power. Retain the compact\\
    \hspace*{1em}priority or comparator form\\
    \hspace*{1em}favored by synthesis.};

\end{tikzpicture}%
  }
\caption{Procedural skill types and representative stored skill summaries.
\textit{Flow \& Validation} captures general synthesis and evidence-quality
procedures, \textit{Design Setup} preserves design-specific compilation
knowledge, and \textit{RTL Optimization} records concrete transformations and
their guardrails; dashed boxes show one stored example for each type.}
  \label{fig:procedural-skill-types}
\end{figure}
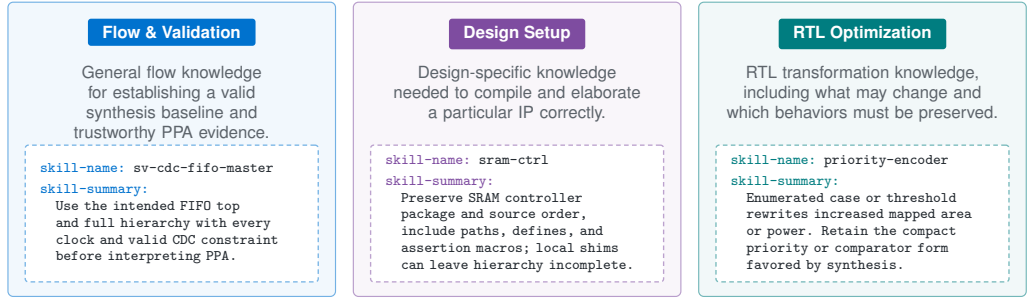

\vspace{-2mm}
\paragraph{Held-out PPA skill retrieval.}
Figure~\ref{fig:procedural-skill-types} shows representative summaries of
skills retrieved during one-shot PPA evaluations of 36 held-out designs with evolving procedural memory. The 50 skill--design retrievals comprised 25 \textit{RTL Optimization}, 21 \textit{Flow \& Validation}, and 4 \textit{Design Setup}  retrievals, indicating that the agent most often received concrete RTL lessons and procedures for obtaining trustworthy
synthesis results. Cohort A produced 38 skill--design retrievals, compared with 12 in cohort B (Appendix~\ref{app:heldout-skill-inv}). Beyond its larger size, cohort A contains more FIFOs, arbiters, and peripheral controllers similar to designs represented in the skill library, whereas cohort B contains more heterogeneous processor, memory, and high-speed interconnect blocks.

\begin{table}[H]
\centering
\footnotesize
\setlength{\tabcolsep}{7pt}
\renewcommand{\arraystretch}{1.0}
\caption{Held-out transfer on ten unseen tasks per category, evaluated once per setting with frozen, read-only memory.}
\label{tab:cvdp-simulation-results}
\begin{tabular*}{\textwidth}{@{\extracolsep{\fill}}lccc@{}}
\toprule
\rowcolor{gray!20}
\multicolumn{4}{c}{\textit{Held-out transfer (frozen, read-only memory)}} \\
\midrule
\textbf{Held-out task category} & \textbf{\edaagent} & \textbf{\chipmemagent} & \textbf{Change} \\
\midrule
CID012 stimulus & \textbf{10/10 (100\%)} & \textbf{10/10 (100\%)} & 0 pp \\
CID013 checker  & 8/10 (80\%) & \textbf{10/10 (100\%)} & \textbf{+20 pp} \\
\midrule
Overall & 18/20 (90\%) & \textbf{20/20 (100\%)} & \textbf{+10 pp} \\
\bottomrule
\end{tabular*}
\end{table}

\FloatBarrier
\paragraph{Held-out CVDP transfer.}
Table~\ref{tab:cvdp-simulation-results} evaluates frozen memory on unseen tasks. CID012 preserves 10/10 performance in both settings. CID013 increases from 8/10 to 10/10 and raises the overall result from 18/20 to 20/20. Because the library is read-only, the two additional passes reflect transfer from previously learned procedural memory. The result shows that stored experience improves checker generation while preserving perfect stimulus generation.

\FloatBarrier


\paragraph{What the memory learns.}
The held-out retrievals show that ChipMEM preserves cross-design RTL knowledge,
not only instructions for operating EDA tools. This distinction matters:
completing the synthesis flow establishes that a result can be evaluated,
whereas improving PPA requires transferable design reasoning. Future work
should isolate the contribution of individual retrieved skills to flow
completion and PPA improvement through paired, skill-level ablations.


\section{Limitations}

ChipMEM uses a fixed top two retrieval setting at threshold $0.6$. The component and frozen transfer studies cover five RTL-OPT designs and ten unseen CVDP tasks per category. Future work should test broader retrieval settings, additional EDA tasks, and controlled runtime environments.

\section{Conclusion}

ChipMEM converts tool verified EDA experience into reusable procedural and statistical memory. It improves accepted outcomes without changing the underlying agent. In the OpenTitan evaluation, it used fewer LLM and tool calls, fewer input and total tokens, and less tool time on average. The SWE-bench Pro \citep{swebenchpro2025} pilot in Appendix~\ref{app:swe-bench-pro} extends the evaluation to software engineering and motivates testing ChipMEM across additional domains. Future work should test larger memories and broader cross-domain transfer.



\bibliographystyle{plainnat}
\bibliography{iclr2026_conference}

\clearpage
\appendix
\thispagestyle{plain}
\begin{center}
    \vspace*{0.3in}
    {\LARGE\bfseries ChipMEM: Verification-Grounded Memory for EDA Agents\par}
    \vspace{0.18in}
    {\Large\bfseries Supplementary Material\par}
\end{center}
\vspace{0.45in}
{\LARGE\bfseries Appendix\par}
\vspace{0.30in}
{\Large\bfseries Table of Contents\par}
\vspace{0.08in}
\hrule
\vspace{0.10in}
\noindent\textcolor{appendixblue}{\textbf{A.1\quad Per-Design PPA Results}}\dotfill\textbf{\pageref{app:per-design-ppa}}\par
\vspace{0.08in}
\noindent\textcolor{appendixblue}{\textbf{A.2\quad Held-Out A and B: Per-Design PPA Results}}\dotfill\textbf{\pageref{app:heldout-ab}}\par
\vspace{0.08in}
\noindent\textcolor{appendixblue}{\textbf{A.3\quad OpenTitan Ten-Design PPA Audit}}\dotfill\textbf{\pageref{app:opentitan10}}\par
\vspace{0.08in}
\noindent\textcolor{appendixblue}{\textbf{A.4\quad Memory ON vs OFF: Cost and Consistency}}\dotfill\textbf{\pageref{app:opentitan-memory-cost}}\par
\vspace{0.08in}
\noindent\textcolor{appendixblue}{\textbf{A.5\quad Execution and Resource Metrics}}\dotfill\textbf{\pageref{app:execution-metrics}}\par
\vspace{0.08in}
\noindent\textcolor{appendixblue}{\textbf{A.6\quad Experiment Grid}}\dotfill\textbf{\pageref{app:experiment-grid}}\par
\vspace{0.08in}
\noindent\textcolor{appendixblue}{\textbf{A.7\quad SWE-bench Pro Pilot}}\dotfill\textbf{\pageref{app:swe-bench-pro}}\par
\vspace{0.08in}
\noindent\textcolor{appendixblue}{\textbf{A.8\quad Prompts}}\dotfill\textbf{\pageref{app:prompts}}\par
\vspace{0.10in}
\hrule
\clearpage
\section{Appendix}

\begin{table*}[t]
\centering
\footnotesize
\setlength{\tabcolsep}{5pt}
\renewcommand{\arraystretch}{1.1}
\caption{Consolidated audited results on RTL-OPT \citep{lu2026new}, RTLRewriter-Bench \citep{yao2024rtlrewriter}, the custom-design PPA suite, and OpenTitan. Equiv. passed denotes equivalence-passing outputs; for RTLRewriter it also includes unchanged outputs accepted as trivially equivalent, while unproven outputs are excluded. RTL-OPT area wins and golden comparisons are reported among equivalence-passing outputs; its mean area gain averages all 38 designs with zero for invalid or non-improving outputs. RTLRewriter reports area wins over scored cases and geometric-mean cell/wire ratios. custom design suite gains retain only positive equivalence-accepted improvements. OpenTitan reports best-of-five strict-gate outcomes; accepted gains come from each design's highest-scoring full-gate attempt, and WNS/TNS changes are in ns. RTL-OPT ChipMEM selects the best valid result per design across three result sets. Best values are bold.}
\label{tab:ppa-results}
\label{tab:opentitan10-ppa}

\begin{tabularx}{\textwidth}{@{}l *{5}{>{\centering\arraybackslash}X}@{\hspace{5pt}}}
\toprule
\rowcolor{gray!20}
\multicolumn{6}{c}{\textit{(a) RTL-OPT}} \\
\midrule
\textbf{Model} & \textbf{Syntax} & \textbf{Equiv. passed} & \shortstack{\textbf{Area $<$}\\\textbf{Suboptimal RTL}\\\textbf{/ equiv. passed}} & \shortstack{\textbf{Area $<$}\\\textbf{Golden RTL}\\\textbf{/ equiv. passed}} & \shortstack{\textbf{Avg. area}\\\textbf{improvement (\%)}} \\
\midrule
\shortstack[l]{GPT-4o\\{\scriptsize\itshape\color{black!65} RTL-OPT}} & \gradecenter{24}{36/38} & \gradecenter{12}{24/38} & \gradecenter{24}{12/24 (50.0\%)} & \gradecenter{30}{\textbf{10/24 (41.7\%)}} & \gradecenter{12}{3.58\%} \\
\shortstack[l]{GPT-4o-mini\\{\scriptsize\itshape\color{black!65} RTL-OPT}} & \gradecenter{22}{35/38} & \gradecenter{12}{24/38} & \gradecenter{12}{7/24 (29.2\%)} & \gradecenter{14}{6/24 (25.0\%)} & \gradecenter{5}{1.69\%} \\
\shortstack[l]{DeepSeek V3\\{\scriptsize\itshape\color{black!65} RTL-OPT}} & \gradecenter{35}{\textbf{38/38}} & \gradecenter{12}{24/38} & \gradecenter{22}{11/24 (45.8\%)} & \gradecenter{30}{\textbf{10/24 (41.7\%)}} & \gradecenter{12}{3.68\%} \\
\shortstack[l]{DeepSeek R1\\{\scriptsize\itshape\color{black!65} RTL-OPT}} & \gradecenter{30}{37/38} & \gradecenter{10}{23/38} & \gradecenter{28}{13/23 (56.5\%)} & \gradecenter{24}{9/23 (39.1\%)} & \gradecenter{18}{4.80\%} \\
\midrule
\shortstack[l]{GPT-5.5\\{\scriptsize\color{black!65}\edaagent}} & \gradecenter{30}{37/38} & \gradecenter{30}{\textbf{33/38}} & \gradecenter{20}{14/33 (42.4\%)} & \gradecenter{18}{10/33 (30.3\%)} & \gradecenter{24}{6.18\%} \\
\shortstack[l]{GPT-5.5$^{*}$\\{\scriptsize\color{black!65}\chipmemagent}} & \gradecenter{30}{37/38} & \gradecenter{30}{\textbf{33/38}} & \gradecenter{32}{\textbf{19/33 (57.6\%)}} & \gradecenter{26}{13/33 (39.4\%)} & \gradecenter{35}{\textbf{8.79\%}} \\
\bottomrule
\end{tabularx}
\vspace{2pt}
\begin{tabularx}{\textwidth}{@{}l l *{4}{>{\centering\arraybackslash}X}@{\hspace{5pt}}}
\rowcolor{gray!20}
\multicolumn{6}{c}{\textit{(b) RTLRewriter-Bench}} \\
\midrule
\textbf{Model} & \textbf{Suite} & \textbf{Equiv. passed} & \textbf{Area wins} & \mbox{\textbf{Avg. area (\%) $\uparrow$}} & \shortstack{\textbf{Cell/Wire ratio}\\\textbf{($\times$ baseline)}} \\
\midrule
\shortstack[l]{GPT-5.5\\{\scriptsize\color{black!65}\edaagent}} & Short & \gradecenter{22}{33/49} & \gradecenter{18}{12/49} & \gradecenter{22}{5.66\%} & \gradecenter{12}{0.927/0.923} \\
\shortstack[l]{GPT-5.5\\{\scriptsize\color{black!65}\chipmemagent}} & Short & \gradecenter{30}{\textbf{37/49}} & \gradecenter{28}{\textbf{15/49}} & \gradecenter{35}{\textbf{8.69\%}} & \gradecenter{22}{\textbf{0.894/0.897}} \\
\midrule
\shortstack[l]{GPT-5.5\\{\scriptsize\color{black!65}\edaagent}} & Long & \gradecenter{12}{2/5} & \gradecenter{10}{1/5} & \gradecenter{25}{6.85\%} & \gradecenter{18}{0.903/0.892} \\
\shortstack[l]{GPT-5.5\\{\scriptsize\color{black!65}\chipmemagent}} & Long & \gradecenter{12}{2/5} & \gradecenter{10}{1/5} & \gradecenter{30}{\textbf{7.47\%}} & \gradecenter{28}{\textbf{0.872/0.858}} \\
\bottomrule
\end{tabularx}
\vspace{2pt}
\begin{tabularx}{\textwidth}{@{}l *{5}{>{\centering\arraybackslash}X}@{\hspace{5pt}}}
\rowcolor{gray!20}
\multicolumn{6}{c}{\textit{(c) custom-design PPA audit}} \\
\midrule
\textbf{Model} & \textbf{Equiv. passed} & \textbf{Area wins} & \mbox{\textbf{Avg. area (\%) $\uparrow$}} & \mbox{\textbf{Avg. power (\%) $\uparrow$}} & \mbox{\textbf{Median timing (\%)}} \\
\midrule
\shortstack[l]{GPT-5.5\\{\scriptsize\color{black!65}\edaagent}} & \gradecenter{18}{8/20} & \gradecenter{18}{3/20} & \gradecenter{18}{0.1400\%} & \gradecenter{18}{0.3879\%} & \gradecenter{24}{\textbf{0.0000\%}} \\
\shortstack[l]{GPT-5.5\\{\scriptsize\color{black!65}\chipmemagent}} & \gradecenter{28}{\textbf{10/20}} & \gradecenter{28}{\textbf{4/20}} & \gradecenter{28}{\textbf{0.3101\%}} & \gradecenter{28}{\textbf{0.5211\%}} & \gradecenter{24}{\textbf{0.0000\%}} \\
\bottomrule
\end{tabularx}
\vspace{2pt}
\begin{tabularx}{\textwidth}{@{}l *{5}{>{\centering\arraybackslash}X}@{\hspace{5pt}}}
\rowcolor{gray!20}
\multicolumn{6}{c}{\textit{(d) OpenTitan ten-design PPA audit}} \\
\midrule
\textbf{Model} & \textbf{Equiv. passed} & \shortstack{\textbf{Full PPA}\\\textbf{gate pass}} & \mbox{\textbf{Avg. area (\%) $\uparrow$}} & \mbox{\textbf{Avg. power (\%) $\uparrow$}} & \shortstack{\textbf{WNS/TNS change}\\\textbf{(ns)}} \\
\midrule
\shortstack[l]{Qwen3.8-27B\\{\scriptsize\color{black!65}\edaagent}} & \gradecenter{24}{\textbf{6/10}} & \gradecenter{18}{1/10} & \gradecenter{18}{0.0519\%} & \gradecenter{5}{0.0000\%} & 0.0/0.0 \\
\shortstack[l]{Qwen3.8-27B\\{\scriptsize\color{black!65}\chipmemagent}} & \gradecenter{24}{\textbf{6/10}} & \gradecenter{32}{\textbf{4/10}} & \gradecenter{28}{\textbf{0.2431\%}} & \gradecenter{35}{\textbf{2.5373\%}} & 0.0/0.0 \\
\bottomrule
\end{tabularx}
\end{table*}
\raggedbottom

\begin{table}[H]
\centering
\footnotesize
\setlength{\tabcolsep}{7pt}
\renewcommand{\arraystretch}{1.0}
\caption{Five-pass CVDP training aggregates with 50 executions per setting.}
\label{tab:cvdp-training-results}
\begin{tabular*}{\textwidth}{@{\extracolsep{\fill}}lccc@{}}
\toprule
\rowcolor{gray!20}
\multicolumn{4}{c}{\textit{Five-pass training aggregate}} \\
\midrule
\textbf{Training task category} & \textbf{\edaagent} & \textbf{\chipmemagent} & \textbf{Change} \\
\midrule
CID012 stimulus & \grade{35}{\textbf{49/50 (98\%)}} & \grade{30}{48/50 (96\%)} & $-2$ pp \\
CID013 checker  & \grade{16}{39/50 (78\%)} & \grade{28}{\textbf{42/50 (84\%)}} & \colorbox{tableblue!20}{\strut\textbf{+6 pp}} \\
\midrule
Overall & \grade{22}{88/100 (88\%)} & \grade{28}{\textbf{90/100 (90\%)}} & \colorbox{tableblue!18}{\strut\textbf{+2 pp}} \\
\bottomrule
\end{tabular*}
\end{table}
\FloatBarrier

\subsection{Per-Design PPA Results}\label{app:per-design-ppa}

\begin{table}[H]
\centering
\footnotesize
\setlength{\tabcolsep}{2.3pt}
\renewcommand{\arraystretch}{0.90}
\caption{Per-design five-attempt PPA results for GPT-5.5 under \edaagent{} and \chipmemagent{}. Pass@$k$ is cumulative full-synthesis completion (\texttt{synth\_status=1.0}) within attempts 1--$k$. Area, power, and timing are positive design-level gains admitted by the documented equivalence check; all other outcomes contribute zero. Equivalence is reported as a binary PASS/FAIL.} 
\label{tab:v3-pool-a-per-design} 

\resizebox{\textwidth}{!}{%
\begin{tabular}{@{}llcccccccccc@{\hspace{\tabcolsep}}}
\toprule
\rowcolor{gray!20}
\multicolumn{12}{c}{\textit{Per-design five-attempt PPA evaluation ($n=20$)}} \\
\midrule
\textbf{Design} & \textbf{Model} & \textbf{Pass@1} & \textbf{Pass@2} & \textbf{Pass@3} & \textbf{Pass@4} & \textbf{Pass@5} & \textbf{Area Gain} & \textbf{Power Gain} & \textbf{Timing Gain} & \textbf{Synthesis} & \textbf{Equiv. pass} \\
\midrule
\multirow{2}{*}{02\_fifo\_sync} & \edaagent & 0 & 0 & 1 & 1 & 1 & 0.0000\% & 0.0000\% & 0.0000\% & 1/5 & FAIL \\
 & \chipmemagent & 0 & 0 & 0 & 0 & 0 & 0.0000\% & 0.0000\% & 0.0000\% & 0/5 & FAIL \\
\cmidrule(lr){1-12}
\multirow{2}{*}{03\_rv32i\_alu} & \edaagent & 1 & 1 & 1 & 1 & 1 & 0.0000\% & 0.0000\% & 0.0000\% & 2/5 & FAIL \\
 & \chipmemagent & 1 & 1 & 1 & 1 & 1 & 0.0000\% & 0.0000\% & 0.0000\% & 1/5 & FAIL \\
\cmidrule(lr){1-12}
\multirow{2}{*}{09\_apb\_gpio} & \edaagent & 0 & 1 & 1 & 1 & 1 & 0.0567\% & 0.3937\% & 0.0000\% & 1/5 & PASS \\
 & \chipmemagent & 0 & 1 & 1 & 1 & 1 & 0.0000\% & 0.3937\% & 0.0000\% & 4/5 & PASS \\
\cmidrule(lr){1-12}
\multirow{2}{*}{11\_arbiter} & \edaagent & 0 & 1 & 1 & 1 & 1 & 0.0000\% & 1.4529\% & 9.5161\% & 2/5 & PASS \\
 & \chipmemagent & 0 & 0 & 1 & 1 & 1 & 0.0000\% & 0.0000\% & 0.0000\% & 3/5 & FAIL \\
\cmidrule(lr){1-12}
\multirow{2}{*}{14\_axi\_adapter\_rd} & \edaagent & 0 & 0 & 0 & 0 & 1 & 0.0000\% & 0.0000\% & 0.0000\% & 1/5 & PASS \\
 & \chipmemagent & 0 & 1 & 1 & 1 & 1 & 0.0000\% & 0.0000\% & 0.0000\% & 2/5 & PASS \\
\cmidrule(lr){1-12}
\multirow{2}{*}{18\_uart8receiver} & \edaagent & 1 & 1 & 1 & 1 & 1 & 0.0000\% & 0.0000\% & 0.0000\% & 2/5 & FAIL \\
 & \chipmemagent & 0 & 0 & 0 & 1 & 1 & 0.0000\% & 0.0000\% & 0.0000\% & 2/5 & FAIL \\
\cmidrule(lr){1-12}
\multirow{2}{*}{ahb3lite\_apb\_bridge} & \edaagent & 0 & 0 & 0 & 1 & 1 & 0.0000\% & 0.0000\% & 0.0000\% & 2/5 & FAIL \\
 & \chipmemagent & 0 & 0 & 0 & 0 & 1 & 2.9366\% & 4.4379\% & 0.0000\% & 1/5 & PASS \\
\cmidrule(lr){1-12}
\multirow{2}{*}{apb\_spi} & \edaagent & 0 & 1 & 1 & 1 & 1 & 0.0000\% & 0.0000\% & 0.0000\% & 2/5 & PASS \\
 & \chipmemagent & 0 & 1 & 1 & 1 & 1 & 0.0000\% & 0.0000\% & 0.0000\% & 2/5 & FAIL \\
\cmidrule(lr){1-12}
\multirow{2}{*}{axi\_lite\_interface} & \edaagent & 1 & 1 & 1 & 1 & 1 & 0.0000\% & 0.0000\% & 0.0000\% & 2/5 & FAIL \\
 & \chipmemagent & 1 & 1 & 1 & 1 & 1 & 0.0000\% & 0.0000\% & 0.0084\% & 3/5 & PASS \\
\cmidrule(lr){1-12}
\multirow{2}{*}{bsg\_ready\_to\_credit\_flow\_converter} & \edaagent & 0 & 0 & 1 & 1 & 1 & 2.6174\% & 5.4759\% & 2.2307\% & 2/5 & PASS \\
 & \chipmemagent & 0 & 0 & 0 & 0 & 0 & 2.6174\% & 5.4759\% & 2.2307\% & 0/5 & PASS \\
\cmidrule(lr){1-12}
\multirow{2}{*}{bsg\_serial\_in\_parallel\_out\_dynamic} & \edaagent & 0 & 0 & 1 & 1 & 1 & 0.0000\% & 0.0000\% & 0.0000\% & 3/5 & PASS \\
 & \chipmemagent & 0 & 0 & 1 & 1 & 1 & 0.0027\% & 0.0943\% & 0.0000\% & 3/5 & PASS \\
\cmidrule(lr){1-12}
\multirow{2}{*}{cdc\_fifo\_master\_hier} & \edaagent & 1 & 1 & 1 & 1 & 1 & 0.0000\% & 0.0000\% & 0.0000\% & 4/5 & FAIL \\
 & \chipmemagent & 0 & 0 & 0 & 1 & 1 & 0.0000\% & 0.0000\% & 0.0000\% & 2/5 & FAIL \\
\cmidrule(lr){1-12}
\multirow{2}{*}{gpio} & \edaagent & 1 & 1 & 1 & 1 & 1 & 0.0000\% & 0.0000\% & 0.0000\% & 1/5 & PASS \\
 & \chipmemagent & 0 & 0 & 0 & 0 & 1 & 0.0000\% & 0.0000\% & 0.0000\% & 1/5 & FAIL \\
\cmidrule(lr){1-12}
\multirow{2}{*}{hmac} & \edaagent & 0 & 0 & 0 & 0 & 0 & 0.0000\% & 0.0000\% & 0.0000\% & 0/5 & FAIL \\
 & \chipmemagent & 0 & 0 & 0 & 0 & 1 & 0.6454\% & 0.0194\% & 0.0000\% & 1/5 & PASS \\
\cmidrule(lr){1-12}
\multirow{2}{*}{i2c} & \edaagent & 0 & 0 & 0 & 0 & 0 & 0.0000\% & 0.0000\% & 0.0000\% & 0/5 & FAIL \\
 & \chipmemagent & 0 & 0 & 0 & 1 & 1 & 0.0000\% & 0.0000\% & 0.0000\% & 1/5 & FAIL \\
\cmidrule(lr){1-12}
\multirow{2}{*}{nand\_flash\_controller} & \edaagent & 0 & 1 & 1 & 1 & 1 & 0.0000\% & 0.0000\% & 0.0000\% & 2/5 & FAIL \\
 & \chipmemagent & 0 & 0 & 0 & 0 & 0 & 0.0000\% & 0.0000\% & 0.0000\% & 0/5 & PASS \\
\cmidrule(lr){1-12}
\multirow{2}{*}{riscv\_fetch} & \edaagent & 0 & 0 & 0 & 1 & 1 & 0.0000\% & 0.0000\% & 0.0000\% & 1/5 & FAIL \\
 & \chipmemagent & 1 & 1 & 1 & 1 & 1 & 0.0000\% & 0.0000\% & 0.0000\% & 4/5 & PASS \\
\cmidrule(lr){1-12}
\multirow{2}{*}{sdram\_axi} & \edaagent & 0 & 0 & 0 & 1 & 1 & 0.1264\% & 0.4348\% & 0.0000\% & 1/5 & PASS \\
 & \chipmemagent & 1 & 1 & 1 & 1 & 1 & 0.0000\% & 0.0000\% & 0.0000\% & 3/5 & PASS \\
\cmidrule(lr){1-12}
\multirow{2}{*}{sram\_ctrl} & \edaagent & 0 & 0 & 0 & 0 & 0 & 0.0000\% & 0.0000\% & 0.0000\% & 0/5 & FAIL \\
 & \chipmemagent & 0 & 1 & 1 & 1 & 1 & 0.0000\% & 0.0000\% & 0.0000\% & 3/5 & FAIL \\
\cmidrule(lr){1-12}
\multirow{2}{*}{sv\_cdc\_fifo\_master} & \edaagent & 1 & 1 & 1 & 1 & 1 & 0.0000\% & 0.0000\% & 0.0000\% & 3/5 & FAIL \\
 & \chipmemagent & 0 & 0 & 0 & 0 & 0 & 0.0000\% & 0.0000\% & 0.0000\% & 0/5 & FAIL \\
\cmidrule(lr){1-12}
\bottomrule
\end{tabular}%
}
\vspace{2pt}
\resizebox{\textwidth}{!}{%
\begin{tabular}{@{}lccccc@{}}
\rowcolor{gray!20}
\multicolumn{6}{c}{\textit{Aggregate cumulative Pass@$k$ (mean $\pm$ std across 20 designs)}} \\
\midrule
\textbf{Model} & \textbf{Pass@1} & \textbf{Pass@2} & \textbf{Pass@3} & \textbf{Pass@4} & \textbf{Pass@5} \\
\midrule
\edaagent & 30.0$\pm$47.0\% & 50.0$\pm$51.3\% & 65.0$\pm$48.9\% & 80.0$\pm$41.0\% & 85.0$\pm$36.6\% \\
\chipmemagent & 20.0$\pm$41.0\% & 40.0$\pm$50.3\% & 50.0$\pm$51.3\% & 65.0$\pm$48.9\% & 80.0$\pm$41.0\% \\
\bottomrule
\end{tabular}%
}
\end{table}
\FloatBarrier

\subsection{Held-Out PPA Cohorts}\label{app:heldout-ab}
We evaluate two disjoint cohorts of previously unseen RTL designs: cohort A
contains 20 designs and cohort B contains 16 designs. Each design is evaluated
once in cohort order. The evaluation begins with procedural skills distilled
from earlier tool-evaluated PPA sessions, while library updates remain enabled.
Consequently, a reused skill may originate from the initial library or from an
earlier run in the ordered held-out evaluation. In the following tables,
\emph{Reused} denotes use of existing skill context, whereas \emph{Mined}
denotes creation of a skill from the current trajectory.

\subsubsection{Per-design outcomes}\label{app:heldout-per-design}

Table~\ref{tab:heldout-ab-per-design} reports the outcome of each held-out
evaluation. The per-design view distinguishes successful synthesis with
available PPA measurements from runs that completed synthesis but did not
produce sufficient metrics for PPA comparison.

\begin{table}[H]
\centering
\scriptsize
\setlength{\tabcolsep}{3.2pt}
\renewcommand{\arraystretch}{0.86}
\caption{Per-design outcomes from the one-shot ChipMEM evaluation of held-out
cohorts A and B using GPT-5.5. Synthesis scores indicate full (1.00) or partial
(0.33) completion. PPA entries are raw relative changes; em dashes indicate
unavailable metrics.}
\label{tab:heldout-ab-per-design}
\resizebox{\textwidth}{!}{%
\begin{tabular}{@{}llccccc@{\hspace{\tabcolsep}}}
\toprule
\rowcolor{gray!20}
\multicolumn{7}{c}{\textit{Held-out A ($n=20$)}} \\
\midrule
\textbf{Design} & \textbf{Memory action} & \textbf{Synthesis} & \textbf{Power} & \textbf{Area} & \textbf{Timing/WNS} & \textbf{PPA available} \\
\midrule
sync\_fifo & Reused & 0.33 & -10.1\% better & -7.2\% better & +13.3\% better & Yes \\
multibit\_fifo\_sync & Reused & 0.33 & --- & --- & --- & No \\
credit\_fifo & Reused & 1.00 & +0.1\% same & +0.1\% same & -6.4\% worse & Yes \\
cdc\_fifo & Reused & 1.00 & --- & --- & --- & No \\
round\_robin\_arbiter & Reused & 1.00 & --- & --- & --- & No \\
snapshot\_arb & Reused & 1.00 & --- & --- & --- & No \\
comparator & Reused & 1.00 & --- & --- & --- & No \\
flow\_reg & Reused & 1.00 & +0.0\% same & +0.0\% same & +0.0\% same & Yes \\
scc & Reused & 0.33 & +0.0\% same & -0.1\% same & +0.0\% same & Yes \\
rob & Reused & 0.33 & -1.6\% better & -5.2\% better & -15.6\% worse & Yes \\
01\_apb\_protocol\_slave & Reused & 1.00 & --- & --- & --- & No \\
10\_async\_fifo & Reused & 0.33 & --- & --- & -100.0\% worse & Yes \\
14\_target\_i2c & Reused & 0.33 & --- & --- & --- & No \\
15\_i2c\_slave & Reused & 0.33 & +1.2\% worse & -0.6\% better & +0.8\% better & Yes \\
16\_ahb\_slave & Reused & 0.33 & --- & --- & --- & No \\
17\_verilog\_spi\_master & Mined & 0.33 & --- & --- & --- & No \\
26\_cache\_controller & Reused & 1.00 & --- & --- & --- & No \\
29\_pwm & Reused & 0.33 & -5.1\% better & -14.5\% better & +0.4\% same & Yes \\
36\_crc & Reused & 1.00 & --- & --- & --- & No \\
40\_lcd\_controller & Reused & 0.33 & --- & --- & --- & No \\
\bottomrule
\end{tabular}%
}
\vspace{2pt}
\resizebox{\textwidth}{!}{%
\begin{tabular}{@{}llccccc@{\hspace{\tabcolsep}}}
\rowcolor{gray!20}
\multicolumn{7}{c}{\textit{Held-out B ($n=16$)}} \\
\midrule
\textbf{Design} & \textbf{Memory action} & \textbf{Synthesis} & \textbf{Power} & \textbf{Area} & \textbf{Timing/WNS} & \textbf{PPA available} \\
\midrule
20\_hazard\_detection\_unit\_master & Reused & 1.00 & --- & --- & --- & No \\
23\_1\_ethernet\_mac\_ip\_64 & Mined & 0.33 & -0.5\% same & --- & --- & Yes \\
23\_2\_ethernet\_mac\_ip & Reused & 1.00 & --- & --- & --- & No \\
24\_1\_pci\_express\_dma\_if\_axi & Mined & 0.33 & --- & -0.1\% same & +0.0\% same & Yes \\
24\_2\_pci\_express\_dma\_if\_pcie & Mined & 0.33 & -1.8\% better & +0.7\% worse & +0.0\% same & Yes \\
27\_1\_risc\_v\_core\_fetch & Reused & 1.00 & --- & --- & --- & No \\
27\_2\_risc\_v\_core\_mmu & Reused & 1.00 & --- & +0.0\% same & +0.0\% same & Yes \\
32\_adc & Mined & 0.33 & --- & +283.3\% worse & -0.1\% same & Yes \\
33\_dac & Reused & 1.00 & --- & --- & --- & No \\
35\_ddr\_controller & Mined & 0.33 & -3.6\% better & +2.3\% worse & -5.0\% worse & Yes \\
45\_fme\_ctrl\_vm & Mined & 0.33 & --- & --- & --- & No \\
apbi2c & Mined & 1.00 & --- & --- & --- & No \\
axi\_master\_design & Reused & 1.00 & --- & --- & --- & No \\
pcie\_controller\_fsm & Mined & 1.00 & --- & --- & --- & No \\
tri\_mode\_ethernet\_mac & Reused & 0.33 & --- & --- & --- & No \\
wb2axi & Mined & 0.33 & --- & --- & --- & No \\
\bottomrule
\end{tabular}%
}
\end{table}

\clearpage
\subsubsection{Skill inventory}\label{app:heldout-skill-inv}

Skills are grouped into three types. \textit{Flow \& Validation} covers tool
setup, source ingestion, synthesis, constraints, and report collection.
\textit{Design Setup} captures compilation and synthesizability requirements
specific to a design family. \textit{RTL Optimization} records concrete RTL
transformations and the conditions under which they produced positive,
neutral, or negative results.

The retrieval columns report the number of designs in each cohort for which
the agent received a skill, either as the selected skill or as additional
context. A skill is counted at most once per design. Retrieval frequency
therefore measures exposure to a skill, not whether it was applied or caused
the resulting PPA outcome.
\vspace{0.5em}
\footnotesize
\begin{longtable}{@{}p{0.23\textwidth}p{0.49\textwidth}>{\centering\arraybackslash}p{0.085\textwidth}>{\centering\arraybackslash}p{0.085\textwidth}@{}}
\caption{Skill inventory and held-out retrieval frequency.}
\label{tab:skill-inventory-retrieval}\\
\toprule
\textbf{Skill} & \textbf{Skill summary} & \textbf{\shortstack{Cohort A\\retrievals}} & \textbf{\shortstack{Cohort B\\retrievals}} \\
\midrule
\endfirsthead
\multicolumn{4}{c}{\tablename\ \thetable\ (continued)} \\
\toprule
\textbf{Skill} & \textbf{Skill summary} & \textbf{\shortstack{Cohort A\\retrievals}} & \textbf{\shortstack{Cohort B\\retrievals}} \\
\midrule
\endhead
\midrule
\multicolumn{4}{r}{Continued on next page} \\
\endfoot
\bottomrule
\endlastfoot
\rowcolor{gray!20}\multicolumn{4}{@{}l}{\textit{Flow \& Validation}} \\
\addlinespace[2pt]
\texttt{sv-\allowbreak{}cdc-\allowbreak{}fifo-\allowbreak{}master} & Require the intended CDC FIFO top, complete hierarchy, all clocks, and valid CDC constraints to synthesize before comparing area, power, or slack. & 7 & 0 \\
\texttt{02-\allowbreak{}fifo-\allowbreak{}sync} & Fix the source list and elaborate the intended FIFO top before changing RTL. Use the baseline only after synthesis produces timing, area, and power reports. & 4 & 0 \\
\texttt{18-\allowbreak{}uart8receiver} & Count the UART receiver run only after the intended top elaborates, maps, and produces timing, area, and power reports. & 3 & 0 \\
\texttt{ahb3lite-\allowbreak{}apb-\allowbreak{}bridge} & A frontend warning does not necessarily mean the bridge failed. Check that the AHB-to-APB top and its hierarchy elaborate before stopping the run. & 2 & 0 \\
\texttt{nand-\allowbreak{}flash-\allowbreak{}controller} & A missing-top error is a source-ingestion problem, not a PPA result. Fix the flash-controller filelist and elaboration before changing RTL. & 2 & 0 \\
\texttt{streaming-\allowbreak{}datapath} & Use a complete source list and fixed constraints, then write timing, area, and power reports to known files before comparing datapath revisions. & 0 & 2 \\
\texttt{dma-\allowbreak{}engine} & Resolve the DMA top, complete source list, constraints, and report paths before modifying datapath or control RTL. & 0 & 1 \\
\texttt{03-\allowbreak{}rv32i-\allowbreak{}alu} & Reading the HDL is not enough. The intended processor top must elaborate and synthesize before the ALU has usable QoR data. & 0 & 0 \\
\texttt{apb-\allowbreak{}spi} & If the APB-SPI top is missing after the HDL read, check the exact module name, source list, and child-module coverage before editing RTL. & 0 & 0 \\
\texttt{memory-\allowbreak{}controller} & Use the controller's PPA results only after the top and full hierarchy synthesize under a defined constraint set and the reports contain numeric results. & 0 & 0 \\
\texttt{serial-\allowbreak{}interface-\allowbreak{}controller} & When the serial-controller top is missing after the HDL read, use the parser diagnostics to fix the source list, search path, or language mode. & 0 & 0 \\
\addlinespace[3pt]
\rowcolor{gray!20}\multicolumn{4}{@{}l}{\textit{Design Setup}} \\
\addlinespace[2pt]
\texttt{bsg-\allowbreak{}ready-\allowbreak{}to-\allowbreak{}credit-\allowbreak{}flow-\allowbreak{}converter} & Keep the original source order, include paths, defines, and synthesis guards. Changing the preprocessor context can hide the converter top or select the wrong RTL branch. & 1 & 0 \\
\texttt{gpio} & Supply the assertion macros or synthesis-only stubs expected by the GPIO sources. Do not change functional RTL to work around missing verification headers. & 1 & 0 \\
\texttt{i2c} & Use the intended assertion headers, defines, include directories, and synthesis macro branch. Unresolved macros can prevent the I²C top from parsing. & 1 & 0 \\
\texttt{sram-\allowbreak{}ctrl} & Preserve package and source order, include directories, defines, and assertion macros in the SRAM controller filelist. Local shims can change preprocessing or leave the hierarchy incomplete. & 1 & 0 \\
\texttt{fsm-\allowbreak{}controller} & Code asynchronous reset and synchronous initialization as separate conditions. Only the true asynchronous reset belongs in the reset branch. & 0 & 0 \\
\addlinespace[3pt]
\rowcolor{gray!20}\multicolumn{4}{@{}l}{\textit{RTL Optimization}} \\
\addlinespace[2pt]
\texttt{riscv-\allowbreak{}fetch} & Delete only state proven constant or unobservable. Preserve redirects, stalls, flushes, skid-buffer behavior, PC updates, reset behavior, and cycle latency. & 2 & 3 \\
\texttt{cdc-\allowbreak{}fifo-\allowbreak{}master-\allowbreak{}hier} & Synthesize the complete FIFO hierarchy with no black boxes. Pointer or register-enable rewrites increased mapped cost and must preserve the CDC, flag, and memory behavior. & 3 & 1 \\
\texttt{11-\allowbreak{}arbiter} & Recoding the round-robin state-update logic increased mapped cost. Limit changes to local common-term or decode cleanup, then compare post-synthesis QoR. & 3 & 0 \\
\texttt{axi-\allowbreak{}lite-\allowbreak{}interface} & Explicit register enables, hold-mux rewrites, and idle-data gating were neutral or worse after mapping and can change AXI-Lite-visible behavior. & 3 & 0 \\
\texttt{dma-\allowbreak{}interface} & Derive load enables for wide DMA registers from existing transfer or state conditions. Preserve reset values and interface latency, then measure switching, area, and slack after mapping. & 0 & 3 \\
\texttt{09-\allowbreak{}apb-\allowbreak{}gpio} & Vectorizing per-bit GPIO logic or factoring APB address decode can map worse. Preserve the register map, read/write behavior, interrupts, and cycle timing. & 2 & 0 \\
\texttt{14-\allowbreak{}axi-\allowbreak{}adapter-\allowbreak{}rd} & Parameter-controlled tie-offs on disabled AXI read paths produced no QoR gain, likely because synthesis had already removed the inactive logic. & 2 & 0 \\
\texttt{bsg-\allowbreak{}serial-\allowbreak{}in-\allowbreak{}parallel-\allowbreak{}out-\allowbreak{}dynamic} & A one-entry FIFO controller can be recoded only if depth, valid/occupancy behavior, handshake timing, latency, throughput, and parameter behavior stay unchanged. & 1 & 0 \\
\texttt{hazard-\allowbreak{}detector} & Factor repeated opcode and register-decode terms and use direct equality comparisons in combinational hazard logic. Prove equivalence and compare mapped QoR. & 0 & 1 \\
\texttt{priority-\allowbreak{}encoder} & Replacing a compact priority or comparator cone with enumerated case or threshold logic increased mapped area or power. Keep the RTL form the mapper handles better. & 0 & 1 \\
\texttt{bus-\allowbreak{}bridge} & Forcing combinational bridge signals to zero or adding speculative enables increased logic and hurt slack. Keep gating only when mapped power, area, and timing improve. & 0 & 0 \\
\texttt{hmac} & Build only the digest or message word selected in the current cycle instead of a full-width combinational bus. Recheck timing because the narrower mux cone may have different depth. & 0 & 0 \\
\texttt{protocol-\allowbreak{}bridge} & Remove identical-arm ternaries, redundant self-assignments, combinational self-feedback, and arithmetic identities in the bridge control logic. Resynthesize each change separately. & 0 & 0 \\
\texttt{sdram-\allowbreak{}axi} & Operand or payload gating on AXI/SDRAM data paths added muxing and could raise area or dynamic power. Keep it only when mapped QoR improves under the same activity assumptions. & 0 & 0 \\
\addlinespace[3pt]
\end{longtable}

\FloatBarrier

\subsection{OpenTitan Ten-Design PPA Audit}\label{app:opentitan10}
\begin{table}[H]
\centering
\footnotesize
\setlength{\tabcolsep}{6pt}
\renewcommand{\arraystretch}{1.0}
\caption{Complete per-design cumulative strict-gate audit for the OpenTitan experiment. Attempts 1--5 are complete for all ten designs in both modes. All runs use Qwen3.8-27B and the same frozen harness. A pass requires the strict equivalence and OpenROAD gate to succeed.}
\label{tab:opentitan10-per-design}
\resizebox{\textwidth}{!}{%
\begin{tabular}{@{}llccccc@{}}
\toprule
\rowcolor{gray!20}
\multicolumn{7}{c}{\textit{OpenTitan per-design cumulative binary audit ($n=10$)}} \\
\midrule
\textbf{Design} & \textbf{Model} & \textbf{Pass@1} & \textbf{Pass@2} & \textbf{Pass@3} & \textbf{Pass@4} & \textbf{Pass@5} \\
\midrule
\multirow{2}{*}{aon\_timer} & \edaagent & 1 & 1 & 1 & 1 & 1 \\
 & \chipmemagent & 0 & 0 & 1 & 1 & 1 \\
\cmidrule(lr){1-7}
\multirow{2}{*}{rv\_timer} & \edaagent & 0 & 0 & 0 & 0 & 0 \\
 & \chipmemagent & 0 & 0 & 0 & 0 & 0 \\
\cmidrule(lr){1-7}
\multirow{2}{*}{pattgen} & \edaagent & 0 & 0 & 0 & 0 & 0 \\
 & \chipmemagent & 0 & 0 & 0 & 1 & 1 \\
\cmidrule(lr){1-7}
\multirow{2}{*}{i2c} & \edaagent & 0 & 0 & 0 & 0 & 0 \\
 & \chipmemagent & 0 & 0 & 0 & 0 & 0 \\
\cmidrule(lr){1-7}
\multirow{2}{*}{spi\_host} & \edaagent & 0 & 0 & 0 & 0 & 0 \\
 & \chipmemagent & 0 & 0 & 1 & 1 & 1 \\
\cmidrule(lr){1-7}
\multirow{2}{*}{hmac} & \edaagent & 0 & 0 & 0 & 0 & 0 \\
 & \chipmemagent & 0 & 0 & 0 & 0 & 0 \\
\cmidrule(lr){1-7}
\multirow{2}{*}{dma} & \edaagent & 0 & 0 & 0 & 0 & 0 \\
 & \chipmemagent & 0 & 0 & 0 & 0 & 0 \\
\cmidrule(lr){1-7}
\multirow{2}{*}{keymgr} & \edaagent & 0 & 0 & 0 & 0 & 0 \\
 & \chipmemagent & 0 & 0 & 1 & 1 & 1 \\
\cmidrule(lr){1-7}
\multirow{2}{*}{aes} & \edaagent & 0 & 0 & 0 & 0 & 0 \\
 & \chipmemagent & 0 & 0 & 0 & 0 & 0 \\
\cmidrule(lr){1-7}
\multirow{2}{*}{kmac} & \edaagent & 0 & 0 & 0 & 0 & 0 \\
 & \chipmemagent & 0 & 0 & 0 & 0 & 0 \\
\midrule
\rowcolor{gray!20}
\multicolumn{7}{c}{\textit{Aggregate cumulative Pass@$k$ (mean $\pm$ sample std across ten designs)}} \\
\midrule
\multicolumn{2}{l}{\textbf{\edaagent}} & \textbf{10.0\% $\pm$ 31.6\%} & \textbf{10.0\% $\pm$ 31.6\%} & 10.0\% $\pm$ 31.6\% & 10.0\% $\pm$ 31.6\% & 10.0\% $\pm$ 31.6\% \\
\multicolumn{2}{l}{\chipmemagent} & 0.0\% $\pm$ 0.0\% & 0.0\% $\pm$ 0.0\% & \textbf{30.0\% $\pm$ 48.3\%} & \textbf{40.0\% $\pm$ 51.6\%} & \textbf{40.0\% $\pm$ 51.6\%} \\
\bottomrule
\end{tabular}%
}
\end{table}
\FloatBarrier

\subsection{Memory ON vs OFF---Full Cost, Success, and Consistency Data}\label{app:opentitan-memory-cost}
Figure~\ref{fig:opentitan-memory-efficiency} summarizes the interaction-cost comparison; this section reports the complete task- and attempt-level evidence. The completed legacy run mined a skill after every memory-on session, including all 45 failed sessions, so these results are descriptive of this run rather than a clean evaluation of the current PASS-only learning rule.

\subsubsection*{A.4.1---Per-design success}
\begin{table}[H]
\centering
\small
\caption{Per-design success. EDA Agent + ChipMEM produces passing, equivalence-verified edits on 4/10 designs versus 1/10 for EDA Agent, including three designs (\texttt{pattgen}, \texttt{spi\_host}, and \texttt{keymgr}) on which the EDA Agent fails every attempt.}
\label{tab:appendix-success}
\begin{tabular*}{\textwidth}{@{\extracolsep{\fill}}lccc@{}}
\toprule
\textbf{Design} & \textbf{OFF passes (of 5)} & \textbf{ON passes (of 5)} & \textbf{Note} \\
\midrule
\texttt{aon\_timer} & 2 & 1 & \\
\texttt{pattgen} & 0 & 2 & Unlocked by memory \\
\texttt{spi\_host} & 0 & 1 & Unlocked by memory \\
\texttt{keymgr} & 0 & 1 & Unlocked by memory \\
\texttt{rv\_timer} & 0 & 0 & \\
\texttt{i2c} & 0 & 0 & \\
\texttt{hmac} & 0 & 0 & \\
\texttt{dma} & 0 & 0 & \\
\texttt{aes} & 0 & 0 & \\
\texttt{kmac} & 0 & 0 & \\
\midrule
\textbf{Total} & \textbf{2 passes; 1/10 designs} & \textbf{5 passes; 4/10 designs} & \\
\bottomrule
\end{tabular*}
\end{table}

\subsubsection*{A.4.2---Overall interaction cost}
\begin{table}[H]
\centering
\small
\caption{Overall interaction cost across 50 sessions per mode. Output tokens and session wall time favor EDA Agent; the increase is consistent with memory writes, reflection, and longer-running successful sessions. Wall time is cumulative session duration, not parallel-run makespan.}
\label{tab:appendix-cost}
\begin{tabularx}{\textwidth}{@{}Xrrrrr@{}}
\toprule
\textbf{Metric} & \textbf{OFF total} & \textbf{ON total} & \textbf{OFF mean} & \textbf{ON mean} & \textbf{$\Delta$ mean} \\
\midrule
Total tokens  & 15.357M & 14.834M & 307.1k & 296.7k & $-3.4\%$ \\
Input tokens  & 14.516M & 13.530M & 290.3k & 270.6k & $-6.8\%$ \\
Output tokens & 0.841M  & 1.304M  & 16.8k  & 26.1k  & $+55.0\%$ \\
LLM calls     & 1,468   & 1,303   & 29.36  & 26.06  & $-11.2\%$ \\
Tool calls    & 1,372   & 1,166   & 27.44  & 23.32  & $-15.0\%$ \\
Tool time     & 350.19 min & 308.14 min & 7.00 min & 6.16 min & $-12.0\%$ \\
Wall time     & 2,048.4 min & 2,353.2 min & 41.0 min & 47.1 min & $+14.9\%$ \\
\bottomrule
\end{tabularx}
\vspace{2pt}
\parbox{\textwidth}{\footnotesize\emph{Medians.} LLM calls: OFF 40 (the budget cap), ON 27. Tool calls: OFF 36.5, ON 18. Total tokens: OFF 373.6k, ON 379.2k.}
\end{table}

\subsubsection*{A.4.3---Attempt-level paired comparison}
\begin{table}[H]
\centering
\small
\caption{Pairwise deltas over 50 design-and-repeat matched attempts. Efficiency deltas are directionally favorable to EDA Agent + ChipMEM, but clustered confidence intervals cross zero except for output tokens, which increase with memory. Primary full-gate outcome summaries appear in Table~\ref{tab:appendix-success}.}
\label{tab:appendix-pairwise}
\begin{tabular*}{\textwidth}{@{\extracolsep{\fill}}lrrrl@{}}
\toprule
\textbf{Metric} & \textbf{ON lower} & \textbf{Tied} & \textbf{ON higher} & \textbf{Cluster-bootstrap 95\% CI of mean $\Delta$} \\
\midrule
Input tokens  & 31 & 0  & 19 & $[-58{,}049,\ +13{,}391]$ \\
Output tokens & 18 & 0  & 32 & $[+318,\ +20{,}726]$ \\
Total tokens  & 27 & 0  & 23 & $[-54{,}672,\ +29{,}844]$ \\
LLM calls     & 25 & 20 & 5  & $[-7.04,\ +0.30]$ \\
Tool calls    & 29 & 10 & 11 & $[-8.74,\ +0.16]$ \\
Tool seconds  & 25 & 0  & 25 & $[-141.1,\ +19.6]$ \\
Wall seconds  & 19 & 0  & 31 & $[-196.5,\ +1{,}062.8]$ \\
\bottomrule
\end{tabular*}
\end{table}

\subsubsection*{A.4.4---Per-design consistency}
\begin{table}[H]
\centering
\small
\caption{Direction of the per-design mean change across the ten OpenTitan designs.}
\label{tab:appendix-consistency}
\begin{tabular*}{0.72\textwidth}{@{\extracolsep{\fill}}lccc@{}}
\toprule
\textbf{Metric} & \textbf{ON better} & \textbf{Tie} & \textbf{ON worse} \\
\midrule
Total tokens & 6/10 & 0 & 4/10 \\
LLM calls    & 7/10 & 2/10 & 1/10 \\
Tool calls   & 8/10 & 0 & 2/10 \\
Tool time    & 5/10 & 0 & 5/10 \\
\bottomrule
\end{tabular*}
\vspace{2pt}
\parbox{0.72\textwidth}{\footnotesize\emph{Note.} The LLM-call ties are \texttt{keymgr} and \texttt{aes}; both reach the 40-call cap in both modes.}
\end{table}

\subsubsection*{A.4.6---Matched case study: \texttt{aon\_timer}}
\begin{table}[H]
\centering
\small
\caption{On the only design with at least one full-gate pass in both modes, EDA Agent + ChipMEM uses roughly half the token and call cost and one-third less session wall time. Values are per-attempt means.}
\label{tab:appendix-aontimer}
\begin{tabular*}{0.72\textwidth}{@{\extracolsep{\fill}}lrrr@{}}
\toprule
\textbf{Metric} & \textbf{OFF} & \textbf{ON} & \textbf{$\Delta$} \\
\midrule
Total tokens & 321.5k & 156.3k & $-51\%$ \\
Input tokens & 292.6k & 138.6k & $-53\%$ \\
LLM calls    & 26.6 & 13.6 & $-49\%$ \\
Tool calls   & 24.0 & 12.6 & $-48\%$ \\
Tool time    & 40.2 s & 22.7 s & $-44\%$ \\
Wall time    & 1,303.5 s & 858.0 s & $-34\%$ \\
\bottomrule
\end{tabular*}
\end{table}
\FloatBarrier

\subsection{Execution and Resource Metrics}\label{app:execution-metrics}
\FloatBarrier

\begin{table}[H]
\centering
\scriptsize
\setlength{\tabcolsep}{2.5pt}
\renewcommand{\arraystretch}{1.05}
\caption{Execution audit for the 100 Qwen3.8-27B canonical sessions underlying Table~\ref{tab:opentitan10-ppa} and Appendix Table~\ref{tab:opentitan10-per-design}: ten designs, five attempts, and two modes. Continuous entries are mean $\pm$ sample standard deviation. Across the 50 sessions per mode, \edaagent{}/\chipmemagent{} used 15.357M/14.834M total tokens, 34.14/39.22 cumulative session-hours, 1468/1303 LLM calls, and 1372/1166 tool calls. All canonical rows and transcripts were read from the completed Brev snapshots. Baseline/candidate OpenROAD times are reported only where gate metrics were recovered; their available $n$ is shown. Four source-migrated attempt-1 transcripts do not reconstruct the retained canonical tool-time field exactly, so the table uses the preserved canonical values. The OpenTitan canonical schema did not preserve a retry counter; aborted-session counts were recovered from the exact transcript flags. Cached-token and reasoning-token counts were unavailable and are not reported as zero.}
\label{tab:opentitan10-execution-metrics}
\resizebox{\textwidth}{!}{%
\begin{tabular}{@{}llccccccccc@{}}
\toprule
\rowcolor{gray!20}
\multicolumn{11}{c}{\textit{OpenTitan latency, tokens, and interaction counts}} \\
\midrule
\textbf{Mode} & \textbf{Attempt} & \textbf{$n$} & \textbf{Wall (min)} & \textbf{Tool (min)} & \textbf{CPU (s)} & \textbf{Input kTok} & \textbf{Output kTok} & \textbf{Total kTok} & \textbf{LLM calls} & \textbf{Tool calls} \\
\midrule
\edaagent & 1 & 10 & $66.1\pm64.7$ & $8.87\pm11.31$ & $0.36\pm0.14$ & $287.2\pm155.8$ & $18.0\pm16.9$ & $305.2\pm159.9$ & $29.1\pm14.2$ & $26.9\pm15.1$ \\
\chipmemagent & 1 & 10 & $52.4\pm37.9$ & $6.92\pm8.24$ & $5.23\pm3.23$ & $274.8\pm171.8$ & $17.4\pm23.7$ & $292.2\pm182.2$ & $26.9\pm14.8$ & $25.0\pm15.1$ \\
\edaagent & 2 & 10 & $33.0\pm23.8$ & $5.96\pm7.50$ & $0.19\pm0.11$ & $274.9\pm147.6$ & $14.2\pm8.7$ & $289.1\pm146.1$ & $27.6\pm13.4$ & $26.0\pm14.2$ \\
\chipmemagent & 2 & 10 & $43.4\pm26.2$ & $5.99\pm7.60$ & $4.86\pm2.99$ & $257.2\pm161.5$ & $23.2\pm26.1$ & $280.4\pm175.4$ & $26.3\pm14.9$ & $24.4\pm14.9$ \\
\edaagent & 3 & 10 & $31.3\pm23.1$ & $6.60\pm7.56$ & $0.20\pm0.12$ & $284.2\pm147.4$ & $12.5\pm12.3$ & $296.7\pm147.9$ & $29.7\pm13.3$ & $28.3\pm14.0$ \\
\chipmemagent & 3 & 10 & $46.4\pm30.1$ & $5.93\pm6.65$ & $5.27\pm3.28$ & $270.3\pm150.8$ & $26.8\pm23.2$ & $297.2\pm160.6$ & $26.4\pm13.3$ & $23.4\pm14.4$ \\
\edaagent & 4 & 10 & $39.0\pm27.9$ & $6.47\pm8.56$ & $0.21\pm0.12$ & $314.9\pm156.4$ & $21.7\pm34.3$ & $336.6\pm171.6$ & $31.4\pm13.9$ & $29.1\pm14.1$ \\
\chipmemagent & 4 & 10 & $45.8\pm29.2$ & $5.61\pm6.76$ & $5.68\pm3.41$ & $265.3\pm168.4$ & $28.2\pm28.4$ & $293.5\pm187.5$ & $25.3\pm13.6$ & $22.4\pm13.4$ \\
\edaagent & 5 & 10 & $35.4\pm22.5$ & $7.11\pm8.26$ & $0.22\pm0.14$ & $290.3\pm157.1$ & $17.8\pm19.8$ & $308.1\pm164.5$ & $29.0\pm14.5$ & $26.9\pm15.0$ \\
\chipmemagent & 5 & 10 & $47.2\pm28.9$ & $6.36\pm7.83$ & $6.19\pm3.93$ & $285.3\pm159.5$ & $34.8\pm31.1$ & $320.1\pm173.5$ & $25.4\pm12.2$ & $21.4\pm13.1$ \\
\bottomrule
\end{tabular}%
}
\vspace{0.06in}
\resizebox{\textwidth}{!}{%
\begin{tabular}{@{}llccccccccc@{}}
\toprule
\rowcolor{gray!20}
\multicolumn{11}{c}{\textit{OpenTitan gate coverage and memory-control activity}} \\
\midrule
\textbf{Mode} & \textbf{Attempt} & \textbf{Baseline ORFS (min)} & \textbf{Candidate ORFS (min)} & \textbf{Gate $n$} & \textbf{Equiv. pass} & \textbf{Pass/Fail} & \textbf{Aborted} & \textbf{Bayes updates} & \textbf{Nudges} & \textbf{Retrieved/Mined} \\
\midrule
\edaagent & 1 & $28.9\pm49.9$ & $22.8\pm33.9$ & 8/10 & 5/10 & 1/9 & 6/10 & $0.0\pm0.0$ & $0.0\pm0.0$ & $0.0\pm0.0$ / 0 \\
\chipmemagent & 1 & $38.0\pm56.4$ & $14.9\pm13.4$ & 6/10 & 3/10 & 0/10 & 7/10 & $25.0\pm15.1$ & $1.6\pm1.4$ & $0.0\pm0.0$ / 10 \\
\edaagent & 2 & $32.3\pm48.8$ & $13.6\pm12.9$ & 8/10 & 5/10 & 0/10 & 6/10 & $0.0\pm0.0$ & $0.0\pm0.0$ & $0.0\pm0.0$ / 0 \\
\chipmemagent & 2 & $32.3\pm48.8$ & $13.5\pm12.7$ & 8/10 & 3/10 & 0/10 & 6/10 & $24.4\pm14.9$ & $2.7\pm0.7$ & $0.1\pm0.3$ / 10 \\
\edaagent & 3 & $29.3\pm49.9$ & $11.3\pm13.1$ & 8/10 & 6/10 & 1/9 & 7/10 & $0.0\pm0.0$ & $0.0\pm0.0$ & $0.0\pm0.0$ / 0 \\
\chipmemagent & 3 & $29.0\pm46.7$ & $12.1\pm14.0$ & 9/10 & 6/10 & 3/7 & 4/10 & $23.4\pm14.4$ & $2.7\pm0.7$ & $0.2\pm0.4$ / 10 \\
\edaagent & 4 & $29.0\pm46.7$ & $12.0\pm14.2$ & 9/10 & 6/10 & 0/10 & 6/10 & $0.0\pm0.0$ & $0.0\pm0.0$ & $0.0\pm0.0$ / 0 \\
\chipmemagent & 4 & $32.3\pm48.8$ & $12.4\pm12.5$ & 8/10 & 4/10 & 1/9 & 6/10 & $22.4\pm13.4$ & $2.8\pm0.6$ & $0.2\pm0.4$ / 10 \\
\edaagent & 5 & $32.3\pm48.8$ & $11.8\pm13.4$ & 8/10 & 5/10 & 0/10 & 7/10 & $0.0\pm0.0$ & $0.0\pm0.0$ & $0.0\pm0.0$ / 0 \\
\chipmemagent & 5 & $32.3\pm48.8$ & $10.9\pm12.8$ & 8/10 & 3/10 & 1/9 & 4/10 & $21.4\pm13.1$ & $2.5\pm0.8$ & $0.3\pm0.5$ / 10 \\
\bottomrule
\end{tabular}%
}
\end{table}
\FloatBarrier

\subsection{Experiment Grid}\label{app:experiment-grid}
The main controlled grids contain 200 proprietary PPA executions,
100 OpenTitan executions, and 240 CVDP simulation executions.

\subsection{SWE-bench Pro Pilot}\label{app:swe-bench-pro}

\begin{figure}[H]
\centering
\includegraphics[width=0.58\textwidth]{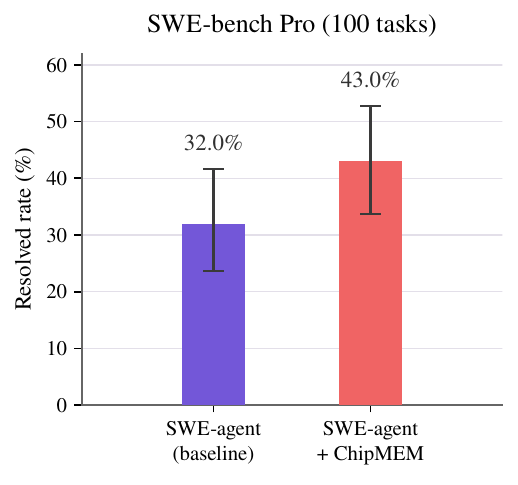}
\caption{SWE-bench Pro resolved rate in a 100-task pilot per mode. Error bars show Wilson 95\% confidence intervals. SWE-agent + ChipMEM resolves 43/100 tasks compared with 32/100 for the SWE-agent baseline, an absolute difference of 11 percentage points.}
\label{fig:swe-bench-pro-pilot}
\end{figure}

\begin{table}[H]
\centering
\small
\caption{Aggregate SWE-bench Pro outcomes for the 100-task pilot. The preserved pilot summary does not include matched per-task outcomes, so we do not report a paired significance test.}
\label{tab:swe-bench-pro-pilot}
\begin{tabular}{lrrrrr}
\toprule
\textbf{Mode} & \textbf{Tasks} & \textbf{Resolved} & \textbf{Unresolved} & \textbf{Rate} & \textbf{95\% CI} \\
\midrule
SWE-agent baseline & 100 & 32 & 68 & 32.0\% & [23.7, 41.7] \\
SWE-agent + ChipMEM & 100 & 43 & 57 & 43.0\% & [33.7, 52.8] \\
\bottomrule
\end{tabular}
\end{table}
\FloatBarrier

\subsection{Prompts}\label{app:prompts}
The prompts are grouped by the model and reported experiment in which they were
used. Task-specific values and session data were inserted at runtime.

\paragraph{EDA agent with GPT-5.5.}
The following distillation prompts produced procedural skills for the reported
GPT-5.5 EDA-agent experiments.

\begin{promptbox}{EDA Agent GPT-5.5 Skill-Distillation Prompt}
You distill one agent run into a reusable learned skill.
Use only the supplied session and measured reward. Return strict JSON:
{
  "name": "generic-lowercase-skill-name",
  "description": "one-line description",
  "body": "Markdown with ## DO and ## AVOID sections",
  "metadata": {"failure_modes": "short summary"}
}
Keep rules generic and evidence-based. A failed or regressed action belongs
in AVOID. Successful measured actions belong in DO. Do not include
instance-specific paths or secrets.
\end{promptbox}

\begin{promptbox}{EDA Agent GPT-5.5 Distillation Input Template}
AGENT TYPE: {{ agent_type }}
INITIAL REQUEST: {{ initial_request }}
REWARD: score={{ reward_score }}, success={{ reward_success }},
        components={{ reward_components }}, failure_class={{ failure_class }}
{% if existing_skill_body %}

EXISTING SKILL TO UPDATE:
{{ existing_skill_body }}
{% endif %}

SESSION:
{{ session_transcript }}
\end{promptbox}

\paragraph{EDA agent with Qwen3.8-27B on OpenTitan.}
The following agent and distillation prompts were used only for the reported
OpenTitan experiment.

\begin{promptbox}{EDA Agent Qwen3.8-27B OpenTitan Prompt}
You are an autonomous engineer. Solve the task using the available tools.
When done, reply with your final summary instead of a tool call.

Optimize OpenTitan IP {{ design }} for lower physical area and power without
changing behavior or ports. You may edit only the files returned by
list_files. First inspect the RTL and run orfs_synth for a baseline. Make
conservative changes, rerun orfs_synth after each change, and keep only
measured improvements. A strict equivalence check and full ORFS flow will
grade the result.
\end{promptbox}

\begin{promptbox}{Qwen3.8-27B OpenTitan Distillation Prompt}
Distill exactly one reusable flat PPA skill from this scored OpenTitan PASS
trajectory. Return only concise DO and AVOID lines.
\end{promptbox}

The measured session trajectory was supplied to the distiller together with
this fixed instruction. Memory OFF and Memory ON used the same agent prompt,
model settings, tools, and execution budget. Memory ON additionally received
the retrieved procedural skills.

\end{document}